\documentclass[11pt]{scrartcl}
\DeclareOldFontCommand{\rm}{\normalfont\rmfamily}{\mathrm}
\DeclareOldFontCommand{\sf}{\normalfont\sffamily}{\mathsf}
\DeclareOldFontCommand{\tt}{\normalfont\ttfamily}{\mathtt}
\DeclareOldFontCommand{\bf}{\normalfont\bfseries}{\mathbf}
\DeclareOldFontCommand{\it}{\normalfont\itshape}{\mathit}
\DeclareOldFontCommand{\sl}{\normalfont\slshape}{\@nomath\sl}
\DeclareOldFontCommand{\sc}{\normalfont\scshape}{\@nomath\sc}
\DeclareRobustCommand*\cal{\@fontswitch\relax\mathcal}
\DeclareRobustCommand*\mit{\@fontswitch\relax\mathnormal}
\usepackage[utf8]{inputenc}
\usepackage{color}
\usepackage{type1cm}
\usepackage{graphicx}
\usepackage[usenames,svgnames]{xcolor}
\usepackage[hyphens]{url}
\usepackage[colorlinks=true,linkcolor=blue,citecolor=blue,breaklinks=true]{hyperref}
\usepackage[numbers,sort&compress]{natbib}

\usepackage{titling}
\predate{}
\postdate{}

\usepackage{amsmath,amssymb,bm,amsthm,cases}
\usepackage{cleveref}
\DeclareMathOperator{\Tr}{\mathrm{Tr}}
\newcommand{\del}{\partial}
\usepackage{braket}

\usepackage{authblk}

\theoremstyle{plain}
\newtheorem{theorem}{Theorem}[section]

\theoremstyle{definition}
\newtheorem{definition}[theorem]{Definition}
\newtheorem{assumption}[theorem]{Assumption}
\theoremstyle{remark}

\crefname{assumption}{Assumption}{Assumptions}

\begin{document}
\nocite{apsrev41control}
\title{\LARGE Power-law escape rate of SGD}
\author[1]{\normalsize Takashi Mori}
\author[2]{Liu Ziyin}
\author[2]{Kangqiao Liu}
\author[1,2,3]{Masahito Ueda}
\affil[1]{\it RIKEN Center for Emergent Matter Science (CEMS), Wako, Saitama 351-0198, Japan}
\affil[2]{\it Department of Physics, The University of Tokyo, Bunkyo-ku, Tokyo 113-0033, Japan}
\affil[3]{\it Institute for Physics of Intelligence, The Uninversity of Tokyo, Bunkyo-ku, Tokyo 113-0033, Japan}

\date{}
\maketitle

\begin{abstract}
Stochastic gradient descent (SGD) undergoes complicated multiplicative noise for the mean-square loss.
We use this property of SGD noise to derive a stochastic differential equation (SDE) with simpler additive noise by performing a random time change.
Using this formalism, we show that the log loss barrier $\Delta\log L=\log[L(\theta^s)/L(\theta^*)]$ between a local minimum $\theta^*$ and a saddle $\theta^s$ determines the escape rate of SGD from the local minimum, contrary to the previous results borrowing from physics that the linear loss barrier $\Delta L=L(\theta^s)-L(\theta^*)$ decides the escape rate.
Our escape-rate formula strongly depends on the typical magnitude $h^*$ and the number $n$ of the outlier eigenvalues of the Hessian.
This result explains an empirical fact that SGD prefers flat minima with low effective dimensions, giving an insight into implicit biases of SGD.
\end{abstract}

\section{Introduction}
Deep learning has achieved breakthroughs in various applications in artificial intelligence such as image classification~\citep{Krizhevsky2012,LeCun2015}, speech recognition~\citep{Hinton2012}, natural language processing~\citep{Collobert2008}, and natural sciences~\citep{Iten2020, Bapst2020, Seif2021}.
Such unparalleled success of deep learning hinges crucially on stochastic gradient descent (SGD) and its variants as an efficient training algorithm.

Although the loss landscape is highly nonconvex, the SGD often succeeds in finding a global minimum.
It has been argued that the SGD noise plays a key role in escaping from local minima~\citep{Jastrzebski2017, Wu2018, Wu2020, Zhu2019, Meng2020, Xie2021, Liu2021}.
It has also been suggested that SGD has an implicit bias that is beneficial for generalization.
For example, SGD may help the network to find \emph{flat minima}, which are considered to imply good generalization~\citep{Keskar2017, Hoffer2017, Wu2018}, although it is also pointed out that flatness alone does not ensure good generalization~\citep{Dinh2017}.
How and why does SGD help the network escape from bad local minima and find flat minima?
These questions have been addressed in several works, and it is now recognized that the SGD noise strength and structure significantly affect the efficiency of escape from local minima.
Our work follows this line of research and adds new theoretical perspectives.

In physics and chemistry, escape from a local minimum of the energy landscape due to thermal noise at temperature $T$ is a thoroughly studied fundamental problem~\citep{Kramers1940, Eyring1935}.
When the energy barrier is given by $\Delta E$, the escape rate is proportional to $e^{-\Delta E/T}$, which is known as the Arrhenius law.
By analogy, in machine learning, escape from a local minimum of the loss function is considered to be determined by the loss barrier height $\Delta L=L(\theta^s)-L(\theta^*)$, where $L(\theta)$ denotes the loss function at the network parameters $\theta$, $\theta^*$ stands for a local minimum of $L(\theta)$, and $\theta^s$ denotes a saddle point that separates $\theta^*$ from other minima.
Indeed, if we assume that the SGD noise is uniform and isotropic, which is often assumed in machine-learning literature~\citep{Jastrzebski2017}, the escape rate is proportional to $e^{-\Delta L/D}$, where $D$ denotes the SGD noise strength.

In this paper, we show that inhomogeneity of the SGD noise strength in the parameter space brings about drastic modification for the mean-square loss. 
We show that the escape rate is determined by the \emph{logarithmic} loss barrier height $\Delta\log L=\log L(\theta^s)-\log L(\theta^*)=\log[L(\theta^s)/L(\theta^*)]$.
In other words, the escape rate is determined not by the difference but by the ratio of $L(\theta^s)$ and $L(\theta^*)$.
This result means that even if the loss barrier height $\Delta L$ is the same, minima with smaller values of $L(\theta^*)$ are more stable.

Moreover, given the fact that the eigenvalue spectrum of the Hessian at a minimum consists of a bulk of almost zero eigenvalues and outliers~\citep{Sagun2017, Papyan2019}, our escape-rate formula implies that \emph{SGD prefers flat minima with a low effective dimension}, where the effective dimension is defined as the number of outliers~\citep{MacKay1992} and flatness is measured by a typical magnitude of outlier eigenvalues~\citep{Keskar2017}.
The previous theories~\citep{Jastrzebski2017, Wu2018, Zhu2019, Meng2020, Xie2021, Liu2021} have also successfully explained the fact that SGD prefers flat minima, but not the preference of small effective dimensions.
The logarithmic loss naturally explains the latter, and sheds light on implicit biases of SGD.

\textbf{Main contributions}:
We obtain the following main results:

\begin{itemize}
\item We derive an equation for approximating the SGD noise in Eq.~(\ref{eq:SGD_cov_formula}).
Remarkably, the SGD noise strength in the mean-square loss is shown to be proportional to the loss function, which is experimentally verified in \cref{sec:noise_strength}.
A key ingredient in deriving Eq.~(\ref{eq:SGD_cov_formula}) is the \emph{decoupling approximation} given in Eq.~(\ref{eq:decoupling}).
This is a novel approximate method introduced in our analysis, and hence we experimentally verify it in \cref{sec:decoupling}. 
\item We derive a novel stochastic differential equation (SDE) in Eq.~(\ref{eq:SDE_tau}) via a random time change introduced in Eq.~(\ref{eq:tau}).
Although the original SDE~(\ref{eq:SDE}) has a multiplicative noise, the transformed SDE~(\ref{eq:SDE_tau}) has a simple additive noise with the gradient of the logarithmic loss.
This shows the relevance of the logarithmic loss landscape for understanding SGD.
\item We derive a novel form of SGD escape rate from a local minimum in Eq.~(\ref{eq:escape_multi}).
Remarkably, the escape rate takes the power-law form with respect to the ratio between $L(\theta^*)$ and $L(\theta^s)$.
In \cref{sec:stationary}, we experimentally test the validity of this result for a linear regression and a neural network.
\item We show that the escape rate of SGD crucially depends on the flatness and the effective dimension, which implies that SGD has implicit biases towards flat minima with low effective dimension.
We also show in Eq.~(\ref{eq:critical}) that a local minimum with an effective dimension $n$ greater than a certain critical value $n_c$ becomes unstable.
\end{itemize}

\textbf{Related works}:
The role of the SGD noise structure has been discussed in some previous works~\citep{Zhu2019, Xie2021, Liu2021, Meng2020, Wojtowytsch2021}.
It was pointed out that the anisotropic nature of the SGD noise is important: the SGD noise covariance matrix is aligned with the Hessian of the loss function, which is beneficial for escape from sharp minima~\citep{Zhu2019, Xie2021, Liu2021}.
These previous works, however, do not take the parameter dependence of the SGD noise strength into account, and consequently, escape rates derived there depend exponentially on the loss barrier height, which differs from our formula.

Compared with the anisotropy of the SGD noise, the inhomogeneity of the SGD noise strength has been less explored. 
In \citep{Meng2020, Wojtowytsch2021}, the SGD dynamics under a state-dependent noise is discussed.
However, in these previous works, the connection between the noise strength and the loss function was not theoretically established, and the logarithmic loss landscape was not discussed.
The instability due to large effective dimensions was also not shown.
Another recent work~\citep{Pesme2021} observed that the noise is proportional to the loss for specific simple models.
In our paper, such a result is derived for more generic models.
\citet{Gurbuzbalaban2021} showed that SGD will converge to a heavy-tailed stationary distribution due to a multiplicative nature of the SGD noise in a simple linear regression problem.
Our paper strengthens this result: we argue that such a heavy-tailed distribution generically appears for the mean-square loss.

\section{Background}

\subsection{Setup}

We consider supervised learning.
Let $\mathcal{D}=\{(x^{(\mu)},y^{(\mu)}): \mu=1,2,\dots,N\}$ be the training dataset, where $x^{(\mu)}\in\mathbb{R}^d$ denotes a data vector and $y^{(\mu)}\in\mathbb{R}$ be its label.
The network output for a given input $x$ is denoted by $f(\theta,x)\in\mathbb{R}$, where $\theta\in\mathbb{R}^P$ stands for a set of trainable parameters with $P$ being the number of trainable parameters.
In this work, we focus on the mean-square loss
\begin{equation}
L(\theta)=\frac{1}{2N}\sum_{\mu=1}^N\left[f(\theta,x^{(\mu)})-y^{(\mu)}\right]^2=:\frac{1}{N}\sum_{\mu=1}^N\ell_\mu(\theta).
\end{equation}
The training proceeds through optimization of $L(\theta)$.
In most machine-learning applications, the optimization is done via SGD or its variants.
In SGD, the parameter $\theta_{k+1}$ at the time step $k+1$ is determined by
\begin{equation}
\theta_{k+1}=\theta_k-\eta\nabla L_{B_k}(\theta_k), \quad L_{B_k}(\theta)=\frac{1}{2B}\sum_{\mu\in B_k}\ell_\mu(\theta),
\label{eq:SGD}
\end{equation}
where $\eta>0$ is the learning rate, $B_k\subset\{1,2,\dots,N\}$ with $|B_k|=B$ is a mini-batch used at the $k$th time step, and $L_{B_k}$ denotes the mini-batch loss.
Since the training dataset $\mathcal{D}$ is randomly divided into mini-batches, the dynamics defined by Eq.~(\ref{eq:SGD}) is stochastic.
When $B=N$, the full training data samples are used for every iteration.
In this case, the dynamics is deterministic and called gradient descent (GD).
SGD is interpreted as GD with stochastic noise.
By introducing the SGD noise $\xi_k=-[\nabla L_{B_k}(\theta_k)-\nabla L(\theta_k)]$, Eq.~(\ref{eq:SGD}) is rewritten as
\begin{equation}
\theta_{k+1}=\theta_k-\eta\nabla L(\theta_k)+\eta\xi_k.
\label{eq:SGD_xi}
\end{equation}
Obviously, $\braket{\xi_k}=0$, where the brackets denote the average over possible choices of mini-batches.
The noise covariance matrix is defined as $\Sigma(\theta_k):=\braket{\xi_k\xi_k^\mathrm{T}}$.
The covariance structure of the SGD noise is important in analyzing the SGD dynamics, which will be discussed in \cref{sec:SGD_noise}.

\subsection{Stochastic differential equation for SGD}

When the parameter update for each iteration is small, which is typically the case when the learning rate $\eta$ is sufficiently small, the continuous-time approximation can be used~\citep{Li2017, Smith-Le2018}.
By introducing a continuous time variable $t\in\mathbb{R}$ and regarding $\eta$ as an infinitesimal time step $dt$, we have a SDE
\begin{equation}
d\theta_t=-\nabla L(\theta_t)dt+\sqrt{\eta\Sigma(\theta_t)}\cdot dW_t,
\label{eq:SDE}
\end{equation}
where $dW_t\sim\mathcal{N}(0,I_Pdt)$ with $I_n$ being the $n$-by-$n$ identity matrix, and the multiplicative noise $\sqrt{\eta\Sigma(\theta_t)}\cdot dW_t$ is interpreted as It\^o since the noise $\xi_k$ in Eq.~(\ref{eq:SGD_xi}) should not depend on $\theta_{k+1}$. 
Throughout this work, we consider the continuous-time approximation~(\ref{eq:SDE}) with Gaussian noise.

In machine learning, the gradient Langevin dynamics (GLD) is also considered, in which the isotropic and uniform Gaussian noise is injected into the GD as
\begin{equation}
d\theta_t=-\nabla L(\theta_t)dt+\sqrt{2D}dW_t,
\label{eq:GLD}
\end{equation}
where $D>0$ corresponds to the noise strength (it is also called the diffusion coefficient)~\citep{Sato2014, Zhang2017a, Zhu2019}.
The GLD yields an escape rate proportional to $e^{-\Delta L/D}$~\citep{Eyring1935, Kramers1940}.
We will see in \cref{sec:main} that the SGD noise structure, which is characterized by $\Sigma(\theta)$, drastically alters the escape rate from a local minimum.

\section{Dynamics of SGD}

The fluctuation of SGD and the local landscape are the two most important factors that influence the escaping behavior of learning. 
In \cref{sec:SGD_noise,sec:log_loss}, we establish the form of the noise of SGD and show that under this type of noise, the relevant landscape for consideration should be $\log L$ instead of $L$. We then use these two results to derive the escape rate of SGD in \cref{sec:main}.

\subsection{Structure of the SGD noise covariance}
\label{sec:SGD_noise}

The SGD noise covariance matrix $\Sigma(\theta)$ significantly affects the dynamics~\citep{Jastrzebski2017, Smith-Le2018, Zhu2019, Ziyin2021}.
In this section, under some approximations, we derive the following expression of $\Sigma(\theta)$ for the mean-square loss near a local minimum $\theta^*$:
\begin{equation}
\Sigma(\theta)\approx\frac{2L(\theta)}{B}H(\theta^*),
\label{eq:SGD_cov_formula}
\end{equation}
where $H(\theta)=\nabla^2L(\theta)$ is the Hessian.
It should be noted that $\nabla L(\theta^*)=0$ and $H(\theta^*)$ is positive semidefinite at any local minima $\theta^*$.
We give a derivation below and the list of the approximations and their justifications in Appendix~\ref{appendix:list}.
In particular, the following \emph{decoupling approximation} is a key assumption in the derivation.
\begin{assumption}[decoupling approximation]
The quantities $\ell_\mu$ and 
\[
C_f^{(\mu)}(\theta):=\nabla f(\theta,x^{(\mu)})\nabla f(\theta,x^{(\mu)})^\mathrm{T}
\]
are uncorrelated, which implies
\begin{align}
\frac{1}{N}\sum_{\mu=1}^N\ell_\mu C_f^{(\mu)}(\theta)=\left(\frac{1}{N}\sum_{\mu=1}^N\ell_\mu\right)\cdot\left(\frac{1}{N}\sum_{\mu=1}^NC_f^{(\mu)}\right)
=L(\theta)\frac{1}{N}\sum_{\mu=1}^NC_f^{(\mu)}(\theta).
\label{eq:decoupling}
\end{align}
\end{assumption}
This approximation seems justified for large networks in which $\nabla f$ behaves as a random vector.
In \cref{sec:decoupling}, we experimentally verify the decoupling approximation for the entire training dynamics.

Our formula~(\ref{eq:SGD_cov_formula}) possesses two important properties.
First, the noise is aligned with the Hessian, which is well known and has been pointed out in the literature~\citep{Jastrzebski2017, Zhu2019, Xie2021, Liu2021}.
If the loss landscape has flat directions, which correspond to the directions of the Hessian eigenvectors belonging to vanishingly small eigenvalues, the SGD noise does not arise along these directions.
Consequently, the SGD dynamics is frozen along the flat directions, which effectively reduces the dimension of the parameter space explored by the SGD dynamics.
This reduction plays an important role in the escape efficiency.
Indeed, we will see that the escape rate crucially depends on the effective dimension of a given local minimum.

Second, the noise is proportional to the loss function, which is indeed experimentally confirmed in \cref{sec:noise_strength}.
This property has not been pointed out and not been taken into account in previous studies~\citep{Jastrzebski2017, Zhu2019, Xie2021, Liu2021} and therefore gives new insights into the SGD dynamics.
Indeed, this property allows us to formulate the Langevin equation on the logarithmic loss landscape with simple additive noise as discussed in \cref{sec:log_loss}.
This new formalism yields the power-law escape rate summarized as \cref{theo:main}, and the importance of the effective dimension of local minima for their stability.

It should be noted that although Eq.~(\ref{eq:SGD_cov_formula}) is derived for the mean-square loss, the proportionality between the noise strength and the loss function seems to hold for more general loss functions such as the cross-entropy loss (see \cref{appendix:loss} for the detail).

\textbf{Derivation of Eq.~(\ref{eq:SGD_cov_formula})}:
We start from an analytic expression of $\Sigma(\theta)$, which reads
\begin{align}
\Sigma(\theta)&=\frac{1}{B}\frac{N-B}{N-1}\left(\frac{1}{N}\sum_{\mu=1}^N\nabla\ell_\mu\nabla\ell_\mu^\mathrm{T}-\nabla L\nabla L^\mathrm{T}\right) \nonumber \\
&\simeq\frac{1}{B}\left(\frac{1}{N}\sum_{\mu=1}^N\nabla\ell_\mu\nabla\ell_\mu^\mathrm{T}-\nabla L\nabla L^\mathrm{T}\right),
\label{eq:SGD_cov}
\end{align}
where $B\ll N$ is assumed in the second equality.
The derivation of Eq.~(\ref{eq:SGD_cov}) is found in~\citet{Jastrzebski2017, Smith-Le2018}.
Usually, the gradient noise variance dominates the square of the gradient noise mean, and hence the term $\nabla L\nabla L^\mathrm{T}$ in Eq.~(\ref{eq:SGD_cov}) is negligible.

For the mean-square loss, we have $\nabla\ell_\mu=[f(\theta,x^{(\mu)})-y^{(\mu)}]\nabla f(\theta,x^{(\mu)})$, and hence
\begin{align}
\Sigma(\theta)&\approx\frac{2}{BN}\sum_{\mu=1}^N\ell_\mu\nabla f(\theta,x^{(\mu)})\nabla f(\theta,x^{(\mu)})^\mathrm{T} \nonumber \\
&=\frac{2}{BN}\sum_{\mu=1}^N\ell_\mu C_f^{(\mu)}(\theta)\approx\frac{2L(\theta)}{NB}\sum_{\mu=1}^NC_f^{(\mu)}(\theta),
\label{eq:SGD_cov_MSE}
\end{align}
where the decoupling approximation~(\ref{eq:decoupling}) is used in the last equality.
Equation~(\ref{eq:SGD_cov_MSE}) is directly related to the Hessian of the loss function near a (local or global) minimum.
The Hessian $H(\theta)=\nabla^2L(\theta)$ is written as
\begin{align}
H(\theta)=\frac{1}{N}\sum_{\mu=1}^N\left\{C_f^{(\mu)}(\theta)+\left[f(\theta,x^{(\mu)})-y^{(\mu)}\right]\nabla^2f(\theta,x^{(\mu)})\right\}.
\label{eq:Hessian}
\end{align}
It is shown by \citet{Papyan2018} that the last term of Eq.~(\ref{eq:Hessian}) does not contribute to outliers (i.e. large eigenvalues) of the Hessian.
The dynamics near a local minimum is governed by outliers, and hence we can ignore this term.
At $\theta=\theta^*$, we therefore obtain
\begin{equation}
H(\theta^*)\approx\frac{1}{N}\sum_{\mu=1}^NC_f^{(\mu)}(\theta^*).
\label{eq:Hessian_approx}
\end{equation}
Let us assume $C_f^{(\mu)}(\theta)\approx C_f^{(\mu)}(\theta^*)$ for $\theta$ within the valley of a local minimum $\theta^*$. 
We then obtain the desired expression~(\ref{eq:SGD_cov_formula}) by substituting it into Eq.~(\ref{eq:SGD_cov_MSE}).

\subsection{Logarithmized loss landscape}
\label{sec:log_loss}

Let us consider the It\^o SDE~(\ref{eq:SDE}) with the SGD noise covariance~(\ref{eq:SGD_cov_formula}) near a local minimum $\theta^*$, which is written as
\begin{equation}
d\theta_t=-\nabla L(\theta_t)dt+\sqrt{\frac{2\eta L(\theta_t)}{B}H(\theta^*)}\cdot dW_t.
\label{eq:SDE_t}
\end{equation}
Let us consider a stochastic time $t(\tau)$ for $\tau\geq 0$ as
\begin{equation}
\tau=\int_0^{t(\tau)} dt'\, L(\theta_{t'}),
\label{eq:tau}
\end{equation}
and perform a random time change from $t$ to $\tau$~\citep{Oksendal_text}.
Correspondingly, we introduce the Wiener process $d\tilde{W}_\tau\sim\mathcal{N}(0,I_Pd\tau)$.
Since $d\tau=L(\theta_t)dt$, we have $d\tilde{W}_\tau=\sqrt{L(\theta_t)}\cdot dW_t$.
In terms of the notation $\tilde{\theta}_\tau=\theta_t$, Eq.~(\ref{eq:SDE_t}) is expressed as
\begin{align}
d\tilde{\theta}_\tau&=-\frac{1}{L(\tilde{\theta}_\tau)}\nabla L(\tilde{\theta}_\tau)d\tau+\sqrt{\frac{2\eta H(\theta^*)}{B}}d\tilde{W}_\tau \nonumber \\
&=-\left[\nabla\log L(\tilde{\theta}_\tau)\right]d\tau+\sqrt{\frac{2\eta H(\theta^*)}{B}}d\tilde{W}_\tau.
\label{eq:SDE_tau}
\end{align}

We should note that at a global minimum with $L(\theta)=0$, which is realized in an overparameterized regime~\citep{Zhang2017}, the random time change through Eq.~(\ref{eq:tau}) is ill-defined since $\tau$ is frozen at a finite value once the model reaches a global minimum. 
We can overcome this difficulty by adding an infinitesimal constant $\epsilon>0$ to the loss, which makes the loss function positive without changing the finite-time dynamics like the escape from a local minimum $\theta^*$ with $L(\theta^*)>0$. 

In this way, the Langevin equation on the loss landscape $L(\theta)$ with multiplicative noise is transformed to that on the logarithmic loss landscape $U(\theta)=\log L(\theta)$ with simpler additive noise.
This formulation indicates the importance of considering the logarithmic loss landscape $U(\theta)=\log L(\theta)$.
In the following, we use Eq.~(\ref{eq:SDE_tau}) to discuss the escape efficiency from local minima.

\section{Escape rate from local minima}
\label{sec:main}

Now we present our main result on the escape from a local minimum $\theta^*$.
First, we formally define some key concepts regarding the escape problem.
\begin{definition}[basin of attraction]
For a given local minimum $\theta^*$, its basin of attraction $\mathcal{A}_{\theta^*}$ (or a valley of $\theta^*$) is defined as the set of all the starting points $\theta_0$ such that $\theta_t\to\theta^*$ as $t\to\infty$ when there is no noise.
\end{definition}
\begin{definition}[escape rate]
We denote by $P^*$ the total probability within $\mathcal{A}_{\theta^*}$ and by $\mathcal{J}$ the total flux of the probability current across the boundary of $\mathcal{A}_{\theta^*}$.\footnote{By defining the probability current density $J(\theta,t)$ via the continuity equation $\del P(\theta,t)/\del t=-\nabla J(\theta,t)$ for the probability distribution $P(\theta,t)$ of $\theta$ at time $t$, $\mathcal{J}$ is explicitly given by $\mathcal{J}=\int_{\mathcal{A}_{\theta^*}}d\theta\,\nabla\cdot J(\theta,t)=\int_{\del\mathcal{A}_{\theta^*}}da\cdot J(\theta,t)$, where the second equality follows from Gauss' law. 
Here, $\del\mathcal{A}_{\theta^*}$ denotes the boundary of $\mathcal{A}_{\theta^*}$ and $da$ is a vector representing an infinitesimal element of area of the surface.}
The escape rate $\kappa$ is then defined as $\mathcal{J}/P^*$~\citep{Kramers1940}.
\end{definition}

We now calculate the asymptotic behavior of the escape rate in the weak-noise limit $\eta/B\to +0$, which has been investigated in many fields, including physics~\citep{Kramers1940, Langer1969}, chemistry~\citep{Eyring1935}, stochastic processes~\citep{Bovier2004, Berglund2013}, and machine learning~\citep{Jastrzebski2017, Xie2021}.
In these previous studies, the following two basic assumptions are commonly used, which are expected to hold in the weak-noise limit.
Indeed, the Kramers formula~\citep{Kramers1940}, which is obtained by using \cref{theo:qss,theo:MPEP} below, is rigorously justified~\citep{Bovier2004}.
\begin{assumption}[quasi-stationary approximation]
The parameters $\theta$ obey the quasi-stationary distribution near $\theta^*$, which is identified as the stationary distribution restricted to $\mathcal{A}_{\theta^*}$~\citep{Bianchi2016}.
\label{theo:qss}
\end{assumption}
\begin{assumption}[most probable escape path]
The escape from $\theta^*$ is dominated by the most probable escape path (MPEP)~\citep{Freidlin_text, Maier1993}.
It is known that the direction of the MPEP at one point must be the direction of one eigenvector of the Hessian at that point, and the MPEP passes a saddle point $\theta^s$ at the boundary of $\mathcal{A}_{\theta^*}$.
\label{theo:MPEP}
\end{assumption}

Now we introduce further key assumptions in our analysis for $P>1$.
\begin{assumption}[Hessian outliers]
The eigenvalue spectrum of the Hessian $H(\theta^*)$ at $\theta^*$ has $n$ nonzero eigenvalues, which are called ``outliers''.
The remaining $P-n$ eigenvalues are vanishingly small. 
\label{theo:outliers}
\end{assumption}
\begin{assumption}[low-dimensional subspace of SGD]
Outlier eigenvectors of the Hessian $v_1, v_2,\dots, v_n\in\mathbb{R}^P$ do not change within $\mathcal{A}_{\theta^*}$, and the SGD dynamics is restricted to the $n$-dimensional subspace spanned by those outlier eigenvectors ($n$ is called the effective dimension of the local minimum).
\label{theo:subspace}
\end{assumption}
\cref{theo:outliers} is empirically confirmed~\citep{Sagun2017, Papyan2019}. 
Since the SGD dynamics is frozen along flat directions as we pointed out in \cref{sec:SGD_noise}, the restriction to the $n$-dimensional outlier subspace in \cref{theo:subspace} is justified.
Indeed, it is empirically observed that the SGD dynamics is actually restricted to a low-dimensional subspace spanned by top eigenvectors of the Hessian~\citep{Gur-Ari2018}.

Now we are ready to state our main result in the form of the following theorem.
\begin{theorem}
Let the model parameter $\theta_t$ evolve according to the SDE in Eq.~(\ref{eq:SDE_t}).
Under \cref{theo:qss,theo:MPEP,theo:outliers,theo:subspace}, the escape rate $\kappa$ asymptotically behaves as
\begin{equation}
\kappa\sim\frac{\sqrt{h_e^*|h_e^s|}}{2\pi}\left[\frac{L(\theta^s)}{L(\theta^*)}\right]^{-\left(\frac{B}{\eta h_e^*}+1-\frac{n}{2}\right)}
\label{eq:escape_multi}
\end{equation}
as $\eta/B\to+0$\footnote{Here, the notation ``$f(x)\sim g(x)$ as $x\to +0$'' means $\lim_{x\to +0}(f(x)/g(x))=1$.}, where $\theta^s$ is the saddle on the MPEP, $h_e^*$ is the eigenvalue of the Hessian at $\theta^*$ along the MPEP, and $h_e^s$ is the negative eigenvalue of the Hessian at $\theta^s$ along the MPEP.
\label{theo:main}
\end{theorem}
\begin{proof}
The proof is given in Appendix~\ref{sec:proof}.
\end{proof}

From Eq.~(\ref{eq:escape_multi}) we can obtain some implications.
The factor $[L(\theta^s)/L(\theta^*)]^{-\left(\frac{B}{\eta h_e^*}+1-\frac{n}{2}\right)}$ increases with $h_e^*$ and $n$, which indicates that sharp minima (i.e. minima with large $h_e^*$) or minima with large $n$ are unstable.
This fact explains why SGD finds flat minima with a low effective dimension $n$.
Equation~(\ref{eq:escape_multi}) also implies that the effective dimension of any stable minima must satisfy
\begin{equation}
n<n_c:=2\left(\frac{B}{\eta h_e^*}+1\right).
\label{eq:critical}
\end{equation}
The instability due to a large effective dimension is a new insight naturally explained by the picture of the logarithmic loss landscape.

Let us summarize novel aspects of the escape rate derived here, which are not found in previous studies~\citep{Jastrzebski2017, Wu2018, Zhu2019, Meng2020, Xie2021, Liu2021}.
First, the escape rate in Eq.~(\ref{eq:escape_multi}) takes a power-law form with respect to the ratio between $L(\theta^*)$ and $L(\theta^s)$, which differs from a conventional exponential form with respect to the difference $\Delta L=L(\theta^s)-L(\theta^*)$.
Intuitively, escaping behavior must be a result of noise in the gradient, and if there is no noise, gradient descent cannot escape any basin of attraction. 
Our result agrees with this intuition: at the global minimum where $L(\theta^*)=0$, there is no noise, and the SGD must be stuck in where it is with no escaping behavior, which is directly reflected by our formula. However, the standard exponential escape rate formula predicts a non-zero escape rate even when $L(\theta^*)=0$, which cannot be correct in principle.
Second, Eqs.~(\ref{eq:escape_multi}) and (\ref{eq:critical}) imply that the SGD is biased to small effective dimensions $n$.
As we mentioned, it is empirically confirmed that the SGD dynamics is restricted to a low-dimensional subspace spanned by top eigenvectors of the Hessian~\citep{Gur-Ari2018}.
Since the restriction to a low-dimensional subspace greatly reduces the capacity of the network~\citep{Li2018}, Eq.~(\ref{eq:critical}) supports the presence of an implicit regularization via the SGD.

The formula in Eq.~(\ref{eq:escape_multi}) is tested in \cref{sec:stationary}.
Since it is difficult in practice to identify when the barrier crossing occurs, we instead consider the mean first passage time, which imitates the escape time for a non-convex loss landscape.
Let us fix a threshold value of the loss function.
The first passage time $t_p$ is defined as the shortest time at which the loss exceeds the threshold value. 
We identify the threshold value as $L(\theta^s)$, i.e., the loss at the saddle in the escape problem.
It is expected that $t_p$ is similar to the escape time and proportional to $\kappa^{-1}$.
Indeed, at least when noise is isotropic and uniform as in the GLD~(\ref{eq:GLD}), the inverse of the mean first passage time $1/\braket{t_p}$ is logarithmically equivalent to $\kappa$, i.e., $\log\kappa\sim -\log\braket{t_p}$, in the weak-noise asymptotic limit~\citep{Freidlin_text, Berglund2013}.

\section{Experiments}
\label{sec:exp}

Our key theoretical observation is that the SGD noise strength is proportional to the loss function, which is obtained as a result of the decoupling approximation.
This property leads us to the Langevin equation (\ref{eq:SDE_tau}) with the logarithmic loss gradient and an additive noise through a random time change~(\ref{eq:tau}).
Equation~(\ref{eq:SDE_tau}) implies the escape rate~(\ref{eq:escape_multi}).

In \cref{sec:decoupling}, we show that the decoupling approximation is valid during the entire training dynamics.
In \cref{sec:noise_strength}, we measure the SGD noise strength and confirm that it is indeed proportional to the loss function near a minimum.
In \cref{sec:stationary}, we experimentally test the validity of Eq.~(\ref{eq:escape_multi}) for the escape rate.
We will see that numerical results for a linear regression and for a non-linear neural network agree with our theoretical results.

\subsection{Experimental verification of the decoupling approximation}
\label{sec:decoupling}

Let us compare the eigenvalue distribution of the exact matrix $(1/N)\sum_{\mu=1}^N\ell^{(\mu)}C_f^{(\mu)}$ with that of the decoupled one $L(\theta)\cdot(1/N)\sum_{\mu=1}^NC_f^{(\mu)}$ with $C_f^{(\mu)}=\nabla f(\theta,x^{(\mu)})\nabla f(\theta,x^{(\mu)})^\mathrm{T}$.
We consider a binary classification problem using the first $10^4$ samples of the MNIST dataset such that we classify each image into even (its label is $y=+1$) or odd number (its label is $y=-1$).
The network has two hidden layers, each of which has 100 units and the ReLU activation, followed by the output layer of a single unit with no activation.
Starting from the Glorot initialization, the training is performed via SGD with the mean-square loss, where we fix $\eta=0.01$ and $B=100$.
 
\begin{figure*}[bt]
\centering
\begin{tabular}{ccc}
\small{(a) at initialization}&\small{(b) at 50 epochs}&\small{(c) at 500 epochs}\\
\includegraphics[width=0.3\linewidth]{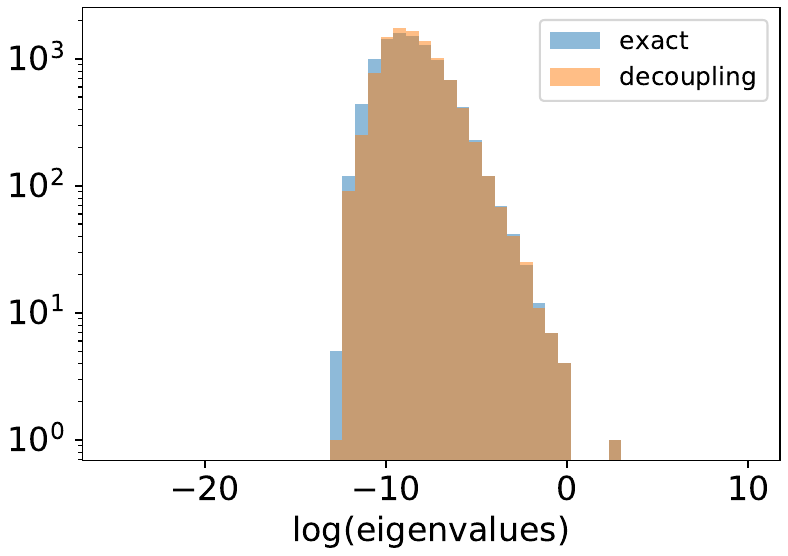}&
\includegraphics[width=0.3\linewidth]{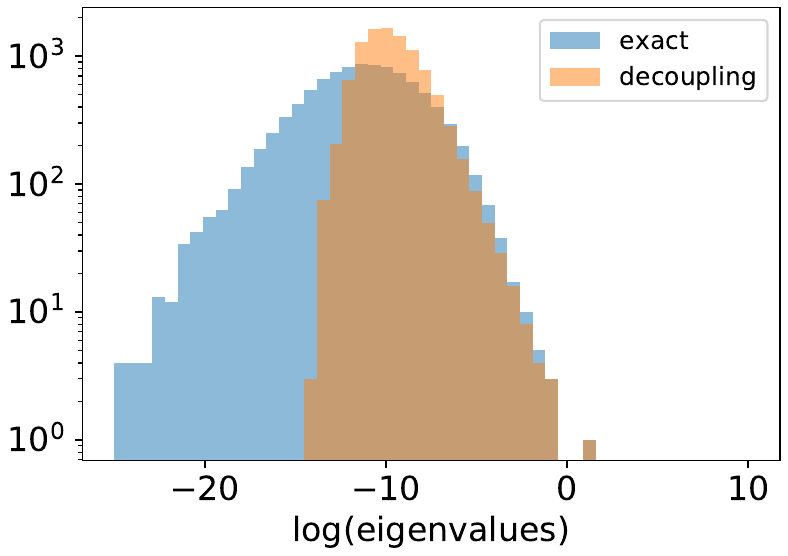}&
\includegraphics[width=0.3\linewidth]{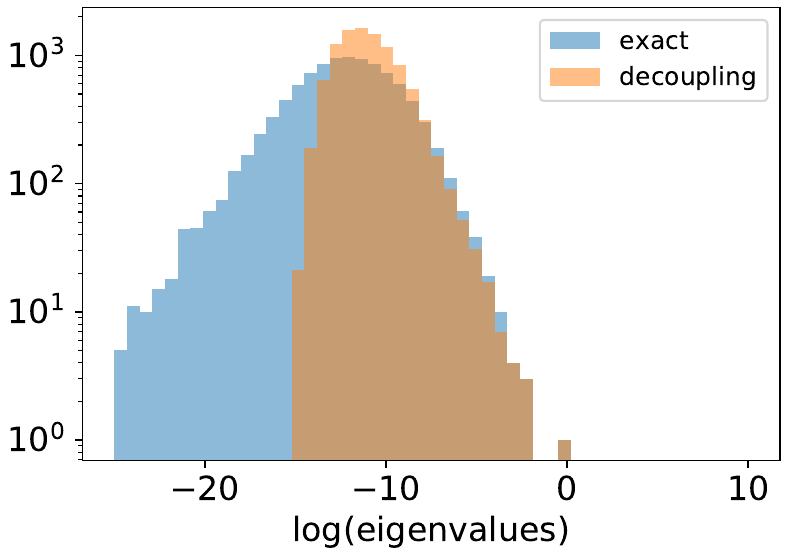}
\end{tabular}
\caption{\small Comparison of the eigenvalue distributions of the left-hand side (exact expression) and the right-hand side (decoupled one) of Eq.~(12) in the main text.
They agree with each other except for exponentially small eigenvalues during the entire training dynamics.}
\label{fig:eigenvalue_Glorot}
\end{figure*}

Figure~\ref{fig:eigenvalue_Glorot} shows histograms of their eigenvalues at different stages of the training: (a) at initialization, (b) after 50 epochs, and (c) after 500 epochs.
We see that the exact matrix and the approximate one have statistically similar eigenvalue distributions except for exponentially small eigenvalues during the training dynamics. 
This shows that the decoupling approximation holds during the entire training dynamics in this experiment.

\subsection{Measurements of the SGD noise strength}
\label{sec:noise_strength}

\begin{figure*}[tb]
\centering
\begin{tabular}{ccc}
\includegraphics[width=0.3\linewidth]{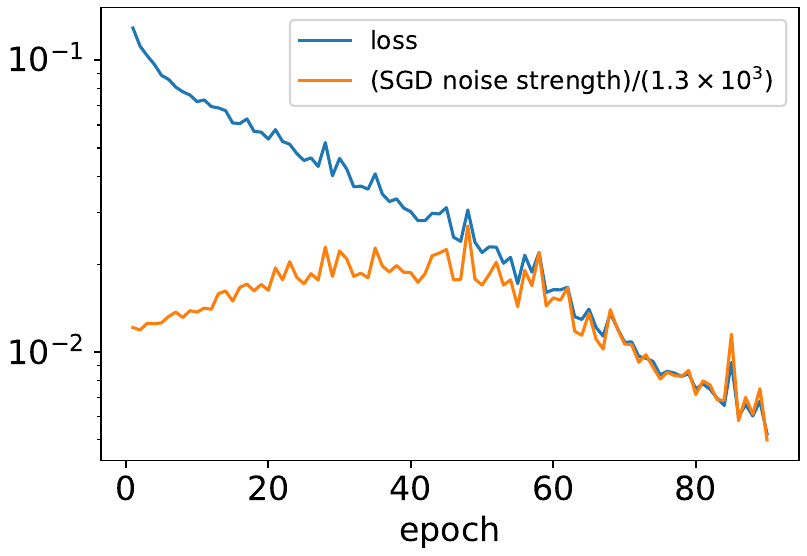}&
\includegraphics[width=0.3\linewidth]{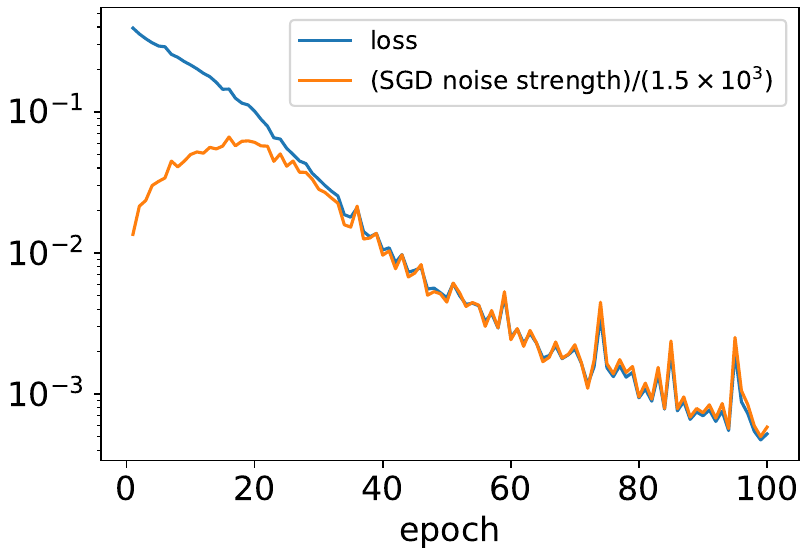}&
\includegraphics[width=0.3\linewidth]{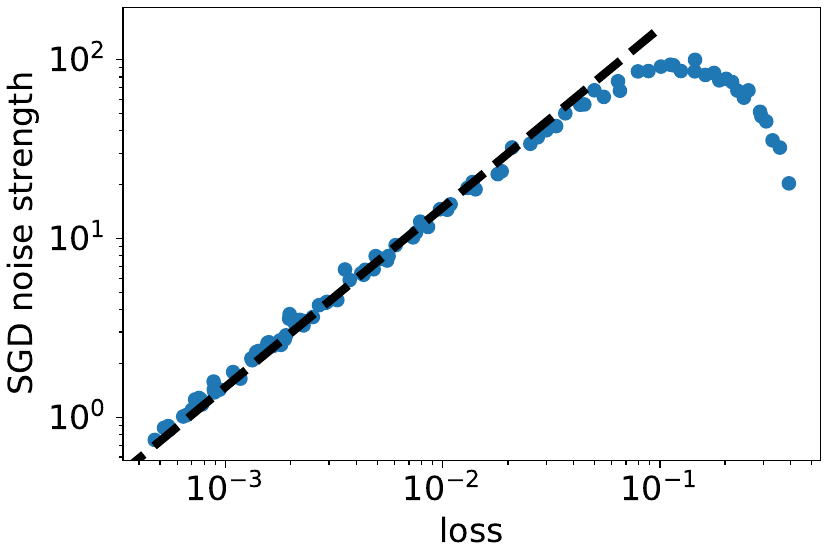}
\end{tabular}
\caption{\small Training dynamics of the loss and the SGD noise strength $\mathcal{N}$.
In the figure, we multiply $\mathcal{N}$ by a numerical factor to emphasize that $\mathcal{N}$ is actually proportional to the loss in a later stage of the training.
(Left) Fully-connected network trained by the Fashion-MNIST dataset. 
(Middle) Convolutional network trained by the CIFAR-10 dataset.
(Right) Loss vs $\mathcal{N}$ in the training of the convolutional network.
The dashed line is a straight line of slope 1, which implies $\mathcal{N}\propto L(\theta)$.}
\label{fig:noise}
\end{figure*}

Now we test whether the SGD noise strength is actually proportional to the loss function, which is predicted by Eq.~(\ref{eq:SGD_cov_formula}), an essential theoretical result of ours.
As a measure of the SGD noise strength, let us consider the norm of the noise vector $\xi$ given by
$\braket{\xi^\mathrm{T}\xi}=\Tr\Sigma\approx\mathcal{N}/B$, where $\mathcal{N}:=(1/N)\sum_{\mu=1}^N\nabla\ell_\mu^\mathrm{T}\nabla\ell_\mu-\nabla L^\mathrm{T}\nabla L$.
Here we present experimental results for two architectures and datasets.
First, we consider training of the Fashion-MNIST dataset by using a fully connected neural network with three hidden layers, each of which has $2\times 10^3$ units and the ReLU activation, followed by the output layer of 10 units with no activation (classification labels are given in the one-hot representation).
Second, we consider training of the CIFAR-10 dataset by using a convolutional neural network.
Following \citet{Keskar2017}, let us denote a stack of $n$ convolutional layers of $a$ filters and a kernel size of $b\times c$ with the stride length of $d$ by $n\times[a,b,c,d]$.
We use the configuration: $3\times[64,3,3,1]$, $3\times[128,3,3,1]$, $3\times[256,3,3,1]$, where a MaxPool(2) is applied after each stack.
To all layers, the ReLU activation is applied.
Finally, an output layer consists of 10 units with no activation.

Starting from the Glorot initialization, the network is trained by SGD of the mini-batch size $B=100$ and $\eta=0.1$ for the mean-square loss.
During the training, we measure the training loss and the noise strength $\mathcal{N}$ for every epoch.
Numerical results are given in Fig.~\ref{fig:noise}.
We see that roughly $\mathcal{N}\propto L$ at a later stage of the training, which agrees with our theoretical prediction.

Although $\mathcal{N}$ is not proportional to $L$ at an early stage of training, it does not mean that Eq.~(\ref{eq:SGD_cov_MSE}) is invalid there.
Since the decoupling approximation is valid for the entire training dynamics, Eq.~(\ref{eq:SGD_cov_MSE}) always holds.
The reason why the SGD noise strength does not decrease with the loss function in the early-stage dynamics is that $\mathcal{N}\approx 2L(\theta)\times (1/N)\sum_{\mu=1}^N\nabla f(\theta,x^{(\mu)})^\mathrm{T}\nabla f(\theta,x^{(\mu)})$, but the quantity $(1/N)\sum_{\mu=1}^N\nabla f(\theta,x^{(\mu)})^\mathrm{T}\nabla f(\theta,x^{(\mu)})$ also changes during training. 

Although Eq.~(\ref{eq:SGD_cov_MSE}) is derived for the mean-square loss, the relation $\mathcal{N}\propto L(\theta)$ holds in more general loss functions; see Appendix~\ref{appendix:loss} for general argument and experiments on the cross entropy loss.

\subsection{Experimental test of the escape rate formula}
\label{sec:stationary}

We now experimentally verify our escape-rate formula~(\ref{eq:escape_multi}).
Below, we first present numerical results for a simple linear regression problem, and then for a nonlinear model, i.e., a neural network.

\begin{figure*}[tb]
\centering
\begin{tabular}{ccc}
\small{(a) Exponent $\phi$ in $P_\mathrm{s}(\theta)$} & \small{(b) $t_p$ for the linear regression} & \small{(c) $t_p$ for the neural network}\\
\includegraphics[width=0.3\linewidth]{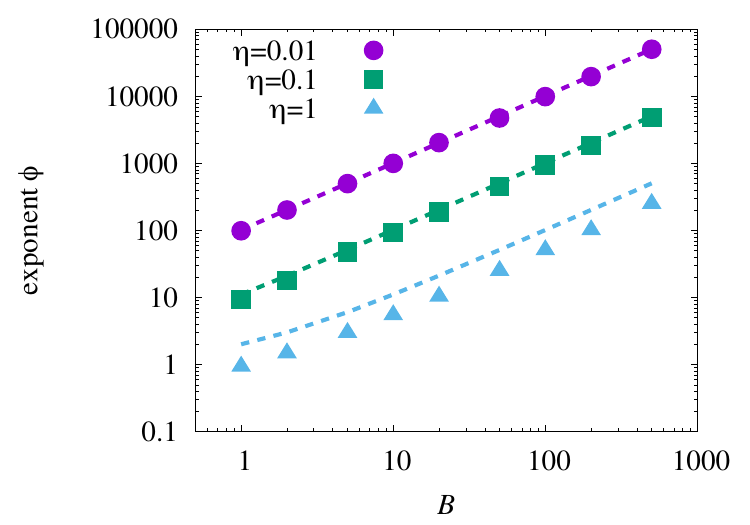}&\hspace{-0.5cm}
\includegraphics[width=0.3\linewidth]{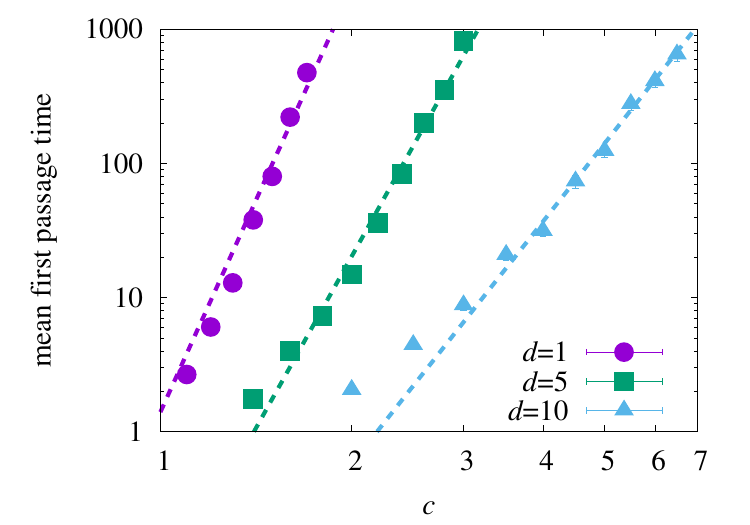}&\hspace{-0.5cm}
\includegraphics[width=0.3\linewidth]{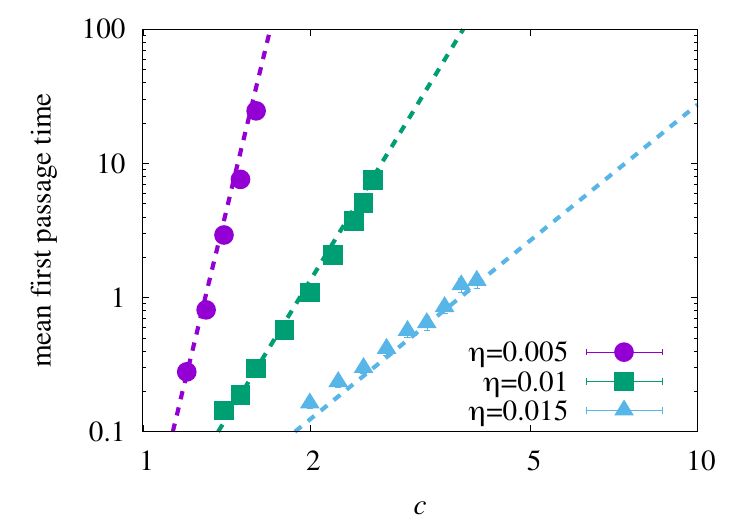}
\end{tabular}
\caption{\small (a) Exponent $\phi$ for the stationary distribution $P_\mathrm{s}(\theta)\propto L(\theta)^{-\phi}$ for $d=1$ in the linear regression. Dashed lines show the theoretical prediction $\phi=1+B/(\eta h^*)$.
(b) Log-log plot of the mean first-passage time $t_p$ vs $c=L(\theta^s)/L(\theta^*)$ for $B=1$ and $\eta=0.1$ in the linear regression.
Dashed lines show the theoretical prediction, $t_p\propto\kappa^{-1}\propto c^{\phi-d/2}$.
(c) Log-log plot of $t_p$ vs $c$ for $B=10$ and various $\eta$ in the neural network.
Dashed lines show the theoretical prediction, $t_p\propto\kappa^{-1}\propto c^{B/(\eta h_e^*)+1-n/2}$ with $h_e^*=95.6$ and $n=9$.}
\label{fig:exponent}
\end{figure*}

\subsubsection{Linear regression}
Let us start with the following linear regression problem: each entry of $x^{(\mu)}\in\mathbb{R}^d$ and its label $y^{(\mu)}\in\mathbb{R}$ are i.i.d. Gaussian random variables of zero mean and unit variance.
We focus on the case of $d\ll N$.
The output for an input $x$ is given by $f(\theta,x)=\theta^\mathrm{T}x$, where $\theta\in\mathbb{R}^d$ is the trainable network parameter.
We optimize $\theta$ via SGD.
The mean-square loss $L(\theta)=(1/2N)\sum_{\mu=1}^N\left(\theta x^{(\mu)}-y^{(\mu)}\right)^2$ is quadratic and has a unique minimum at $\theta\approx 0$.
the Hessian $H=(1/N)\sum_{\mu=1}^Nx^{(\mu)}x^{(\mu)\mathrm{T}}$ has $d$ nonzero eigenvalues, all of which are close to unity.
We can therefore identify $h_e^*=h^*=1$ and $n=d$.

Before investigating the escape rate, we remark that the stationary distribution of $\theta$ in this case is theoretically obtained because of the isotropy of SGD noise.
Indeed, $H(\theta^*)$ in Eq.~(\ref{eq:SDE_tau}) can be replaced by $h^*I_d$, where $I_d$ is the $d$-dimensional identity matrix, and then the stationary distribution of $\tilde{\theta}_\tau$ is simply given by the Gibbs distribution under the potential $U(\theta)=\log L(\theta)$: $\tilde{P}_\mathrm{s}(\theta)\propto e^{-U(\theta)/T}=L(\theta)^{-1/T}$, where the ``temperature'' $T$ is given by $T=\eta h^*/B$.
In Appendix~\ref{appendix:stationary}, it is shown that the stationary distribution $P_\mathrm{s}(\theta)$ of $\theta_t$ under Eq.~(\ref{eq:SDE_t}) is related to $\tilde{P}_\mathrm{s}(\theta)$ via $P_\mathrm{s}(\theta)\propto L(\theta)^{-1}\tilde{P}_\mathrm{s}(\theta)$.
We therefore have
\begin{equation}
P_\mathrm{s}(\theta)\propto L(\theta)^{-\phi}, \quad \phi=1+\frac{B}{\eta h^*}.
\label{eq:stationary}
\end{equation}
Remarkably, it depends on $L(\theta)$ polynomially rather than exponentially as in the standard GLD~\citep{Jastrzebski2017, Sato2014, Zhang2017a}.
This property is closely related to the power-law form of the escape rate.

First, we test Eq.~(\ref{eq:stationary}), i.e. the stationary distribution, for $d=1$ and $N=10^5$.
We sampled the value of $\theta_k$ at every 100 iterations ($k=j\times 100$, $j=1,2,\dots,10^4$) and made a histogram.
We then fit the histogram to the form $P_\mathrm{s}(\theta)\propto L(\theta)^{-\phi}$ and determine the exponent $\phi$.
Our theory predicts $\phi=1+B/(\eta h^*)$.
Numerical results for the exponent $\phi$ are presented in Fig.~\ref{fig:exponent} (a) against $B$ for three fixed learning rates $\eta$.
In the same figure, theoretical values of $\phi$ are plotted in dashed lines. 
The agreement between theory and experiment is fairly well.
For a large learning rate $\eta=1$, the exponent slightly deviates from its theoretical value.
This is due to the effect of a finite learning rate (recall that $\eta$ is assumed to be small in deriving the continuous-time stochastic differential equation).

Next, we test our formula on the escape rate, Eq.~(\ref{eq:escape_multi}).
Although the mean-square loss is quadratic and no barrier crossing occurs, we can measure the first passage time, which imitates the escape time for a non-convex loss landscape as is mentioned in \cref{sec:main}.

The mean first passage time over 100 independent runs is measured for varying threshold values which are specified by $c=L(\theta^s)/L(\theta^*)>1$.
Experimental results for $N=10^4$ are presented in Fig.~\ref{fig:exponent} (b).
Dashed straight lines have slope $B/(\eta h^*)+1-n/2$.
Experiments show that the first passage time behaves as $t_p\propto [L(\theta^s)/L(\theta^*)]^{B/(\eta h^*)+1-n/2}$, which agrees with our theoretical evaluation of $\kappa^{-1}$ [see Eq.~(\ref{eq:escape_multi})].
We conclude that the escape rate crucially depends on the effective dimension $n$, which is not explained by the previous results~\citep{Zhu2019, Xie2021, Liu2021, Meng2020}.

\subsubsection{Neural network}
The escape-rate formula~(\ref{eq:escape_multi}) is also verified in a non-linear model.
As in \cref{sec:decoupling}, we consider the binary classification problem using the first $10^4$ samples of MNINST dataset such that we classify each image into even or odd.
The network has one hidden layer with 10 units activated by ReLU, followed by the output layer of a single unit with no activation.
We always use the mean-square loss.
Starting from the Glorot initialization, the network is pre-trained via SGD with $\eta=0.01$ and $B=100$ for $10^5$ iterations.
We find that after pre-training, the loss becomes almost stationary around at $L(\theta)\approx 0.035$.
We assume that the pre-trained network is near a local minimum.
We then further train the pre-trained network via SGD with a new choice of $\eta$ and $B$ (here we fix $B=10$), and measure the first passage time $t_p$.
The mean first-passage time over 100 independent runs is measured for varying threshold values which are specified by $c=L(\theta^s)/L(\theta^*)>1$.
Experimental results are presented in Fig.~\ref{fig:exponent} (c).
We see the power-law behavior, which is consistent with our theory.

To further verify our theoretical formula~(\ref{eq:escape_multi}), we also compare the power-law exponent for the mean first-passage time with our theoretical prediction $B/(\eta h_e^*)+1-n/2$.
Here, $h_e^*$ and $n$ are estimated by the Hessian eigenvalues.
In Appendix~\ref{sec:Hessian}, we present a numerical result for eigenvalues of the approximate Hessian given by the right-hand side of Eq.~(\ref{eq:Hessian_approx}).
By identifying the largest eigenvalue of the Hessian as $h_e^*$, we have $h_e^*\approx 95.6$. 
On the other hand, it is difficult to precisely determine the effective dimension $n$, but it seems reasonable to estimate $n\sim 10$.
It turns out that theory and experiment agree with each other by choosing $n=9$.
Dashed lines in Fig.~\ref{fig:exponent} (c) correspond to our theoretical prediction $\kappa^{-1}\propto c^{B/(\eta h_e^*)+1-n/2}$ with $h_e^*=95.6$ and $n=9$. 
This excellent agreement shows that our theoretical formula~(\ref{eq:escape_multi}) is also valid in non-linear models.

\section{Conclusion}
In this work, we have investigated the SGD dynamics via a Langevin approach.
With several approximations listed in Appendix~\ref{appendix:list}, we have derived Eq.~(\ref{eq:SGD_cov_formula}), which shows that the SGD noise strength is proportional to the loss function.
This SGD noise covariance structure yields the stochastic differential equation~(\ref{eq:SDE_tau}) with additive noise near a minimum via a random time change~(\ref{eq:tau}).
The original multiplicative noise is reduced to simpler additive noise, but instead the gradient of the loss function is replaced by that of the logarithmic loss function $U(\theta)=\log L(\theta)$.
This new formalism yields the power-law escape rate formula~(\ref{eq:escape_multi}) whose exponent depends on $\eta$, $B$, $h_e^*$, and $n$.

The escape-rate formula in Eq.~(\ref{eq:escape_multi}) explains an empirical fact that SGD favors flat minima with low effective dimensions.
The effective dimension of a minimum must satisfy Eq.~(\ref{eq:critical}) for its stability.
This result as well as the formulation of the SGD dynamics using the logarithmic loss landscape should help understand more deeply the SGD dynamics and its implicit biases in machine learning problems.

Although the present work focuses on the Gaussian noise, the non-Gaussianity can also play an important role.
For example, \citet{Simsekli2019} approximated SGD as a L\'evy-driven SDE, which explains why SGD finds wide minima.
It would be an interesting future problem to take the non-Gaussian effect into account.

\bibliography{apsrevcontrol, deep_learning, physics}

\begin{thebibliography}{42}%
\makeatletter
\providecommand \@ifxundefined [1]{%
 \@ifx{#1\undefined}
}%
\providecommand \@ifnum [1]{%
 \ifnum #1\expandafter \@firstoftwo
 \else \expandafter \@secondoftwo
 \fi
}%
\providecommand \@ifx [1]{%
 \ifx #1\expandafter \@firstoftwo
 \else \expandafter \@secondoftwo
 \fi
}%
\providecommand \natexlab [1]{#1}%
\providecommand \enquote  [1]{``#1''}%
\providecommand \bibnamefont  [1]{#1}%
\providecommand \bibfnamefont [1]{#1}%
\providecommand \citenamefont [1]{#1}%
\providecommand \href@noop [0]{\@secondoftwo}%
\providecommand \href [0]{\begingroup \@sanitize@url \@href}%
\providecommand \@href[1]{\@@startlink{#1}\@@href}%
\providecommand \@@href[1]{\endgroup#1\@@endlink}%
\providecommand \@sanitize@url [0]{\catcode `\\12\catcode `\$12\catcode
  `\&12\catcode `\#12\catcode `\^12\catcode `\_12\catcode `\%12\relax}%
\providecommand \@@startlink[1]{}%
\providecommand \@@endlink[0]{}%
\providecommand \url  [0]{\begingroup\@sanitize@url \@url }%
\providecommand \@url [1]{\endgroup\@href {#1}{\urlprefix }}%
\providecommand \urlprefix  [0]{URL }%
\providecommand \Eprint [0]{\href }%
\providecommand \doibase [0]{https://doi.org/}%
\providecommand \selectlanguage [0]{\@gobble}%
\providecommand \bibinfo  [0]{\@secondoftwo}%
\providecommand \bibfield  [0]{\@secondoftwo}%
\providecommand \translation [1]{[#1]}%
\providecommand \BibitemOpen [0]{}%
\providecommand \bibitemStop [0]{}%
\providecommand \bibitemNoStop [0]{.\EOS\space}%
\providecommand \EOS [0]{\spacefactor3000\relax}%
\providecommand \BibitemShut  [1]{\csname bibitem#1\endcsname}%
\let\auto@bib@innerbib\@empty
\bibitem [{\citenamefont {Krizhevsky}\ \emph {et~al.}(2012)\citenamefont
  {Krizhevsky}, \citenamefont {Sutskever},\ and\ \citenamefont
  {Hinton}}]{Krizhevsky2012}%
  \BibitemOpen
  \bibfield  {author} {\bibinfo {author} {\bibfnamefont {Alex}\ \bibnamefont
  {Krizhevsky}}, \bibinfo {author} {\bibfnamefont {Ilya}\ \bibnamefont
  {Sutskever}},\ and\ \bibinfo {author} {\bibfnamefont {Geoffrey~E}\
  \bibnamefont {Hinton}},\ }\bibfield  {title} {\bibinfo {title} {{Imagenet
  classification with deep convolutional neural networks}},\ }in\ \href
  {http://papers.nips.cc/paper/4824-imagenet-classification-with-deep-convolutional-neural-networks.pdf}
  {\emph {\bibinfo {booktitle} {Advances in Neural Information Processing
  Systems}}}\ (\bibinfo {year} {2012})\BibitemShut {NoStop}%
\bibitem [{\citenamefont {LeCun}\ \emph {et~al.}(2015)\citenamefont {LeCun},
  \citenamefont {Bengio},\ and\ \citenamefont {Hinton}}]{LeCun2015}%
  \BibitemOpen
  \bibfield  {author} {\bibinfo {author} {\bibfnamefont {Yann}\ \bibnamefont
  {LeCun}}, \bibinfo {author} {\bibfnamefont {Yoshua}\ \bibnamefont {Bengio}},\
  and\ \bibinfo {author} {\bibfnamefont {Geoffrey}\ \bibnamefont {Hinton}},\
  }\bibfield  {title} {\bibinfo {title} {{Deep learning}},\ }\href
  {https://doi.org/10.1038/nature14539} {\bibfield  {journal} {\bibinfo
  {journal} {Nature}\ }\textbf {\bibinfo {volume} {521}},\ \bibinfo {pages}
  {436--444} (\bibinfo {year} {2015})}\BibitemShut {NoStop}%
\bibitem [{\citenamefont {Hinton}\ \emph {et~al.}(2012)\citenamefont {Hinton},
  \citenamefont {Deng}, \citenamefont {Yu}, \citenamefont {Dahl}, \citenamefont
  {Mohamed}, \citenamefont {Jaitly}, \citenamefont {Senior}, \citenamefont
  {Vanhoucke}, \citenamefont {Nguyen}, \citenamefont {Sainath},\ and\
  \citenamefont {Others}}]{Hinton2012}%
  \BibitemOpen
  \bibfield  {author} {\bibinfo {author} {\bibfnamefont {Geoffrey}\
  \bibnamefont {Hinton}}, \bibinfo {author} {\bibfnamefont {Li}~\bibnamefont
  {Deng}}, \bibinfo {author} {\bibfnamefont {Dong}\ \bibnamefont {Yu}},
  \bibinfo {author} {\bibfnamefont {George~E}\ \bibnamefont {Dahl}}, \bibinfo
  {author} {\bibfnamefont {Abdel-rahman}\ \bibnamefont {Mohamed}}, \bibinfo
  {author} {\bibfnamefont {Navdeep}\ \bibnamefont {Jaitly}}, \bibinfo {author}
  {\bibfnamefont {Andrew}\ \bibnamefont {Senior}}, \bibinfo {author}
  {\bibfnamefont {Vincent}\ \bibnamefont {Vanhoucke}}, \bibinfo {author}
  {\bibfnamefont {Patrick}\ \bibnamefont {Nguyen}}, \bibinfo {author}
  {\bibfnamefont {Tara~N}\ \bibnamefont {Sainath}},\ and\ \bibinfo {author}
  {\bibnamefont {Others}},\ }\bibfield  {title} {\bibinfo {title} {{Deep neural
  networks for acoustic modeling in speech recognition: The shared views of
  four research groups}},\ }\href {https://doi.org/10.1109/MSP.2012.2205597}
  {\bibfield  {journal} {\bibinfo  {journal} {IEEE Signal processing magazine}\
  }\textbf {\bibinfo {volume} {29}},\ \bibinfo {pages} {82--97} (\bibinfo
  {year} {2012})}\BibitemShut {NoStop}%
\bibitem [{\citenamefont {Collobert}\ and\ \citenamefont
  {Weston}(2008)}]{Collobert2008}%
  \BibitemOpen
  \bibfield  {author} {\bibinfo {author} {\bibfnamefont {Ronan}\ \bibnamefont
  {Collobert}}\ and\ \bibinfo {author} {\bibfnamefont {Jason}\ \bibnamefont
  {Weston}},\ }\bibfield  {title} {\bibinfo {title} {{A unified architecture
  for natural language processing: Deep neural networks with multitask
  learning}},\ }in\ \href {https://doi.org/10.1145/1390156.1390177} {\emph
  {\bibinfo {booktitle} {International Conference on Machine Learning}}}\
  (\bibinfo {year} {2008})\BibitemShut {NoStop}%
\bibitem [{\citenamefont {Iten}\ \emph {et~al.}(2020)\citenamefont {Iten},
  \citenamefont {Metger}, \citenamefont {Wilming}, \citenamefont {{Del Rio}},\
  and\ \citenamefont {Renner}}]{Iten2020}%
  \BibitemOpen
  \bibfield  {author} {\bibinfo {author} {\bibfnamefont {Raban}\ \bibnamefont
  {Iten}}, \bibinfo {author} {\bibfnamefont {Tony}\ \bibnamefont {Metger}},
  \bibinfo {author} {\bibfnamefont {Henrik}\ \bibnamefont {Wilming}}, \bibinfo
  {author} {\bibfnamefont {L{\'{i}}dia}\ \bibnamefont {{Del Rio}}},\ and\
  \bibinfo {author} {\bibfnamefont {Renato}\ \bibnamefont {Renner}},\
  }\bibfield  {title} {\bibinfo {title} {{Discovering Physical Concepts with
  Neural Networks}},\ }\href {https://doi.org/10.1103/PhysRevLett.124.010508}
  {\bibfield  {journal} {\bibinfo  {journal} {Physical Review Letters}\
  }\textbf {\bibinfo {volume} {124}},\ \bibinfo {pages} {10508} (\bibinfo
  {year} {2020})}\BibitemShut {NoStop}%
\bibitem [{\citenamefont {Bapst}\ \emph {et~al.}(2020)\citenamefont {Bapst},
  \citenamefont {Keck}, \citenamefont {Grabska-Barwi{\'{n}}ska}, \citenamefont
  {Donner}, \citenamefont {Cubuk}, \citenamefont {Schoenholz}, \citenamefont
  {Obika}, \citenamefont {Nelson}, \citenamefont {Back}, \citenamefont
  {Hassabis},\ and\ \citenamefont {Kohli}}]{Bapst2020}%
  \BibitemOpen
  \bibfield  {author} {\bibinfo {author} {\bibfnamefont {V.}~\bibnamefont
  {Bapst}}, \bibinfo {author} {\bibfnamefont {T.}~\bibnamefont {Keck}},
  \bibinfo {author} {\bibfnamefont {A.}~\bibnamefont
  {Grabska-Barwi{\'{n}}ska}}, \bibinfo {author} {\bibfnamefont
  {C.}~\bibnamefont {Donner}}, \bibinfo {author} {\bibfnamefont {E.~D.}\
  \bibnamefont {Cubuk}}, \bibinfo {author} {\bibfnamefont {S.~S.}\ \bibnamefont
  {Schoenholz}}, \bibinfo {author} {\bibfnamefont {A.}~\bibnamefont {Obika}},
  \bibinfo {author} {\bibfnamefont {A.~W.R.}\ \bibnamefont {Nelson}}, \bibinfo
  {author} {\bibfnamefont {T.}~\bibnamefont {Back}}, \bibinfo {author}
  {\bibfnamefont {D.}~\bibnamefont {Hassabis}},\ and\ \bibinfo {author}
  {\bibfnamefont {P.}~\bibnamefont {Kohli}},\ }\bibfield  {title} {\bibinfo
  {title} {{Unveiling the predictive power of static structure in glassy
  systems}},\ }\href {https://doi.org/10.1038/s41567-020-0842-8} {\bibfield
  {journal} {\bibinfo  {journal} {Nature Physics}\ }\textbf {\bibinfo {volume}
  {16}},\ \bibinfo {pages} {448--454} (\bibinfo {year} {2020})}\BibitemShut
  {NoStop}%
\bibitem [{\citenamefont {Seif}\ \emph {et~al.}(2021)\citenamefont {Seif},
  \citenamefont {Hafezi},\ and\ \citenamefont {Jarzynski}}]{Seif2021}%
  \BibitemOpen
  \bibfield  {author} {\bibinfo {author} {\bibfnamefont {Alireza}\ \bibnamefont
  {Seif}}, \bibinfo {author} {\bibfnamefont {Mohammad}\ \bibnamefont
  {Hafezi}},\ and\ \bibinfo {author} {\bibfnamefont {Christopher}\ \bibnamefont
  {Jarzynski}},\ }\bibfield  {title} {\bibinfo {title} {{Machine learning the
  thermodynamic arrow of time}},\ }\href
  {https://doi.org/10.1038/s41567-020-1018-2} {\bibfield  {journal} {\bibinfo
  {journal} {Nature Physics}\ }\textbf {\bibinfo {volume} {17}},\ \bibinfo
  {pages} {105--113} (\bibinfo {year} {2021})}\BibitemShut {NoStop}%
\bibitem [{\citenamefont {Jastrz{\c{e}}bski}\ \emph {et~al.}(2017)\citenamefont
  {Jastrz{\c{e}}bski}, \citenamefont {Kenton}, \citenamefont {Arpit},
  \citenamefont {Ballas}, \citenamefont {Fischer}, \citenamefont {Bengio},\
  and\ \citenamefont {Storkey}}]{Jastrzebski2017}%
  \BibitemOpen
  \bibfield  {author} {\bibinfo {author} {\bibfnamefont {Stanis{\l}aw}\
  \bibnamefont {Jastrz{\c{e}}bski}}, \bibinfo {author} {\bibfnamefont
  {Zachary}\ \bibnamefont {Kenton}}, \bibinfo {author} {\bibfnamefont
  {Devansh}\ \bibnamefont {Arpit}}, \bibinfo {author} {\bibfnamefont {Nicolas}\
  \bibnamefont {Ballas}}, \bibinfo {author} {\bibfnamefont {Asja}\ \bibnamefont
  {Fischer}}, \bibinfo {author} {\bibfnamefont {Yoshua}\ \bibnamefont
  {Bengio}},\ and\ \bibinfo {author} {\bibfnamefont {Amos}\ \bibnamefont
  {Storkey}},\ }\bibfield  {title} {\bibinfo {title} {{Three Factors
  Influencing Minima in SGD}},\ }\href {https://arxiv.org/abs/1711.04623}
  {\bibfield  {journal} {\bibinfo  {journal} {arXiv:1711.04623}\ } (\bibinfo
  {year} {2017})}\BibitemShut {NoStop}%
\bibitem [{\citenamefont {Wu}\ \emph {et~al.}(2018)\citenamefont {Wu},
  \citenamefont {Ma},\ and\ \citenamefont {Weinan}}]{Wu2018}%
  \BibitemOpen
  \bibfield  {author} {\bibinfo {author} {\bibfnamefont {Lei}\ \bibnamefont
  {Wu}}, \bibinfo {author} {\bibfnamefont {Chao}\ \bibnamefont {Ma}},\ and\
  \bibinfo {author} {\bibfnamefont {E.}~\bibnamefont {Weinan}},\ }\bibfield
  {title} {\bibinfo {title} {{How SGD selects the global minima in
  over-parameterized learning: A dynamical stability perspective}},\ }in\ \href
  {http://papers.nips.cc/paper/8049-how-sgd-selects-the-global-minima-in-over-parameterized-learning-a-dynamical-stability-perspective}
  {\emph {\bibinfo {booktitle} {Advances in Neural Information Processing
  Systems}}}\ (\bibinfo {year} {2018})\BibitemShut {NoStop}%
\bibitem [{\citenamefont {Wu}\ \emph {et~al.}(2020)\citenamefont {Wu},
  \citenamefont {Hu}, \citenamefont {Xiong}, \citenamefont {Huan},
  \citenamefont {Braverman},\ and\ \citenamefont {Zhu}}]{Wu2020}%
  \BibitemOpen
  \bibfield  {author} {\bibinfo {author} {\bibfnamefont {Jingfeng}\
  \bibnamefont {Wu}}, \bibinfo {author} {\bibfnamefont {Wenqing}\ \bibnamefont
  {Hu}}, \bibinfo {author} {\bibfnamefont {Haoyi}\ \bibnamefont {Xiong}},
  \bibinfo {author} {\bibfnamefont {Jun}\ \bibnamefont {Huan}}, \bibinfo
  {author} {\bibfnamefont {Vladimir}\ \bibnamefont {Braverman}},\ and\ \bibinfo
  {author} {\bibfnamefont {Zhanxing}\ \bibnamefont {Zhu}},\ }\bibfield  {title}
  {\bibinfo {title} {{On the Noisy Gradient Descent that Generalizes as SGD}},\
  }in\ \href {http://arxiv.org/abs/1906.07405} {\emph {\bibinfo {booktitle}
  {International Conference on Machine Learning}}}\ (\bibinfo {year}
  {2020})\BibitemShut {NoStop}%
\bibitem [{\citenamefont {Zhu}\ \emph {et~al.}(2019)\citenamefont {Zhu},
  \citenamefont {Wu}, \citenamefont {Yu}, \citenamefont {Wu},\ and\
  \citenamefont {Ma}}]{Zhu2019}%
  \BibitemOpen
  \bibfield  {author} {\bibinfo {author} {\bibfnamefont {Zhanxing}\
  \bibnamefont {Zhu}}, \bibinfo {author} {\bibfnamefont {Jingfeng}\
  \bibnamefont {Wu}}, \bibinfo {author} {\bibfnamefont {Bing}\ \bibnamefont
  {Yu}}, \bibinfo {author} {\bibfnamefont {Lei}\ \bibnamefont {Wu}},\ and\
  \bibinfo {author} {\bibfnamefont {Jinwen}\ \bibnamefont {Ma}},\ }\bibfield
  {title} {\bibinfo {title} {{The anisotropic noise in stochastic gradient
  descent: Its behavior of escaping from sharp minima and regularization
  effects}},\ }in\ \href {http://proceedings.mlr.press/v97/zhu19e.html} {\emph
  {\bibinfo {booktitle} {International Conference on Machine Learning}}}\
  (\bibinfo {year} {2019})\BibitemShut {NoStop}%
\bibitem [{\citenamefont {Meng}\ \emph {et~al.}(2020)\citenamefont {Meng},
  \citenamefont {Gong}, \citenamefont {Chen}, \citenamefont {Ma},\ and\
  \citenamefont {Liu}}]{Meng2020}%
  \BibitemOpen
  \bibfield  {author} {\bibinfo {author} {\bibfnamefont {Qi}~\bibnamefont
  {Meng}}, \bibinfo {author} {\bibfnamefont {Shiqi}\ \bibnamefont {Gong}},
  \bibinfo {author} {\bibfnamefont {Wei}\ \bibnamefont {Chen}}, \bibinfo
  {author} {\bibfnamefont {Zhi~Ming}\ \bibnamefont {Ma}},\ and\ \bibinfo
  {author} {\bibfnamefont {Tie~Yan}\ \bibnamefont {Liu}},\ }\bibfield  {title}
  {\bibinfo {title} {{Dynamic of Stochastic Gradient Descent with
  State-Dependent Noise}},\ }\href {https://arxiv.org/abs/2006.13719}
  {\bibfield  {journal} {\bibinfo  {journal} {arXiv:2006.13719}\ } (\bibinfo
  {year} {2020})}\BibitemShut {NoStop}%
\bibitem [{\citenamefont {Xie}\ \emph {et~al.}(2021)\citenamefont {Xie},
  \citenamefont {Sato},\ and\ \citenamefont {Sugiyama}}]{Xie2021}%
  \BibitemOpen
  \bibfield  {author} {\bibinfo {author} {\bibfnamefont {Zeke}\ \bibnamefont
  {Xie}}, \bibinfo {author} {\bibfnamefont {Issei}\ \bibnamefont {Sato}},\ and\
  \bibinfo {author} {\bibfnamefont {Masashi}\ \bibnamefont {Sugiyama}},\
  }\bibfield  {title} {\bibinfo {title} {{A Diffusion Theory For Deep Learning
  Dynamics: Stochastic Gradient Descent Exponentially Favors Flat Minima}},\
  }in\ \href {https://openreview.net/forum?id=wXgk_iCiYGo} {\emph {\bibinfo
  {booktitle} {International Conference on Learning Representations}}}\
  (\bibinfo {year} {2021})\BibitemShut {NoStop}%
\bibitem [{\citenamefont {Liu}\ \emph {et~al.}(2021)\citenamefont {Liu},
  \citenamefont {Ziyin},\ and\ \citenamefont {Ueda}}]{Liu2021}%
  \BibitemOpen
  \bibfield  {author} {\bibinfo {author} {\bibfnamefont {Kangqiao}\
  \bibnamefont {Liu}}, \bibinfo {author} {\bibfnamefont {Liu}\ \bibnamefont
  {Ziyin}},\ and\ \bibinfo {author} {\bibfnamefont {Masahito}\ \bibnamefont
  {Ueda}},\ }\bibfield  {title} {\bibinfo {title} {{Noise and Fluctuation of
  Finite Learning Rate Stochastic Gradient Descent}},\ }in\ \href
  {http://arxiv.org/abs/2012.03636} {\emph {\bibinfo {booktitle} {International
  Conference on Machine Learning}}}\ (\bibinfo {year} {2021})\BibitemShut
  {NoStop}%
\bibitem [{\citenamefont {Keskar}\ \emph {et~al.}(2017)\citenamefont {Keskar},
  \citenamefont {Nocedal}, \citenamefont {Tang}, \citenamefont {Mudigere},\
  and\ \citenamefont {Smelyanskiy}}]{Keskar2017}%
  \BibitemOpen
  \bibfield  {author} {\bibinfo {author} {\bibfnamefont {Nitish~Shirish}\
  \bibnamefont {Keskar}}, \bibinfo {author} {\bibfnamefont {Jorge}\
  \bibnamefont {Nocedal}}, \bibinfo {author} {\bibfnamefont {Ping Tak~Peter}\
  \bibnamefont {Tang}}, \bibinfo {author} {\bibfnamefont {Dheevatsa}\
  \bibnamefont {Mudigere}},\ and\ \bibinfo {author} {\bibfnamefont {Mikhail}\
  \bibnamefont {Smelyanskiy}},\ }\bibfield  {title} {\bibinfo {title} {{On
  large-batch training for deep learning: Generalization gap and sharp
  minima}},\ }in\ \href {https://openreview.net/forum?id=H1oyRlYgg} {\emph
  {\bibinfo {booktitle} {International Conference on Learning
  Representations}}}\ (\bibinfo {year} {2017})\BibitemShut {NoStop}%
\bibitem [{\citenamefont {Hoffer}\ \emph {et~al.}(2017)\citenamefont {Hoffer},
  \citenamefont {Hubara},\ and\ \citenamefont {Soudry}}]{Hoffer2017}%
  \BibitemOpen
  \bibfield  {author} {\bibinfo {author} {\bibfnamefont {Elad}\ \bibnamefont
  {Hoffer}}, \bibinfo {author} {\bibfnamefont {Itay}\ \bibnamefont {Hubara}},\
  and\ \bibinfo {author} {\bibfnamefont {Daniel}\ \bibnamefont {Soudry}},\
  }\bibfield  {title} {\bibinfo {title} {{Train longer, generalize better:
  closing the generalization gap in large batch training of neural networks}},\
  }in\ \href {https://doi.org/10.1016/j.jcjd.2014.02.001} {\emph {\bibinfo
  {booktitle} {Advances in Neural Information Processing Systems}}}\ (\bibinfo
  {year} {2017})\BibitemShut {NoStop}%
\bibitem [{\citenamefont {Dinh}\ \emph {et~al.}(2017)\citenamefont {Dinh},
  \citenamefont {Pascanu}, \citenamefont {Bengio},\ and\ \citenamefont
  {Bengio}}]{Dinh2017}%
  \BibitemOpen
  \bibfield  {author} {\bibinfo {author} {\bibfnamefont {Laurent}\ \bibnamefont
  {Dinh}}, \bibinfo {author} {\bibfnamefont {Razvan}\ \bibnamefont {Pascanu}},
  \bibinfo {author} {\bibfnamefont {Samy}\ \bibnamefont {Bengio}},\ and\
  \bibinfo {author} {\bibfnamefont {Yoshua}\ \bibnamefont {Bengio}},\
  }\bibfield  {title} {\bibinfo {title} {{Sharp minima can generalize for deep
  nets}},\ }in\ \href {http://proceedings.mlr.press/v70/dinh17b.html} {\emph
  {\bibinfo {booktitle} {International Conference on Machine Learning}}}\
  (\bibinfo {year} {2017})\BibitemShut {NoStop}%
\bibitem [{\citenamefont {Kramers}(1940)}]{Kramers1940}%
  \BibitemOpen
  \bibfield  {author} {\bibinfo {author} {\bibfnamefont {Hendrik~Anthony}\
  \bibnamefont {Kramers}},\ }\bibfield  {title} {\bibinfo {title} {{Brownian
  motion in a field of force and the diffusion model of chemical reactions}},\
  }\href {https://doi.org/10.1016/S0031-8914(40)90098-2} {\bibfield  {journal}
  {\bibinfo  {journal} {Physica}\ }\textbf {\bibinfo {volume} {7}},\ \bibinfo
  {pages} {284--304} (\bibinfo {year} {1940})}\BibitemShut {NoStop}%
\bibitem [{\citenamefont {Eyring}(1935)}]{Eyring1935}%
  \BibitemOpen
  \bibfield  {author} {\bibinfo {author} {\bibfnamefont {Henry}\ \bibnamefont
  {Eyring}},\ }\bibfield  {title} {\bibinfo {title} {{The Activated Complex in
  Chemical Reactions}},\ }\href {https://doi.org/10.1063/1.1749604} {\bibfield
  {journal} {\bibinfo  {journal} {The Journal of Chemical Physics}\ }\textbf
  {\bibinfo {volume} {3}},\ \bibinfo {pages} {63--71} (\bibinfo {year}
  {1935})}\BibitemShut {NoStop}%
\bibitem [{\citenamefont {Sagun}\ \emph {et~al.}(2017)\citenamefont {Sagun},
  \citenamefont {Evci}, \citenamefont {G{\"{u}}ney}, \citenamefont {Dauphin},\
  and\ \citenamefont {Bottou}}]{Sagun2017}%
  \BibitemOpen
  \bibfield  {author} {\bibinfo {author} {\bibfnamefont {Levent}\ \bibnamefont
  {Sagun}}, \bibinfo {author} {\bibfnamefont {Utku}\ \bibnamefont {Evci}},
  \bibinfo {author} {\bibfnamefont {V.~Ugur}\ \bibnamefont {G{\"{u}}ney}},
  \bibinfo {author} {\bibfnamefont {Yann}\ \bibnamefont {Dauphin}},\ and\
  \bibinfo {author} {\bibfnamefont {L{\'{e}}on}\ \bibnamefont {Bottou}},\
  }\bibfield  {title} {\bibinfo {title} {{Empirical analysis of the hessian of
  over-parametrized neural networks}},\ }\href
  {https://arxiv.org/abs/1706.04454} {\bibfield  {journal} {\bibinfo  {journal}
  {arXiv:1706.04454}\ } (\bibinfo {year} {2017})}\BibitemShut {NoStop}%
\bibitem [{\citenamefont {Papyan}(2019)}]{Papyan2019}%
  \BibitemOpen
  \bibfield  {author} {\bibinfo {author} {\bibfnamefont {Vardan}\ \bibnamefont
  {Papyan}},\ }\bibfield  {title} {\bibinfo {title} {{Measurements of
  three-level hierarchical structure in the outliers in the spectrum of deepnet
  hessians}},\ }in\ \href {http://proceedings.mlr.press/v97/papyan19a.html}
  {\emph {\bibinfo {booktitle} {International Conference on Machine
  Learning}}}\ (\bibinfo {year} {2019})\BibitemShut {NoStop}%
\bibitem [{\citenamefont {MacKay}(1992)}]{MacKay1992}%
  \BibitemOpen
  \bibfield  {author} {\bibinfo {author} {\bibfnamefont {Djc}\ \bibnamefont
  {MacKay}},\ }\bibfield  {title} {\bibinfo {title} {{Bayesian model comparison
  and backprop nets}},\ }in\ \href
  {https://papers.nips.cc/paper/1991/hash/c3c59e5f8b3e9753913f4d435b53c308-Abstract.html}
  {\emph {\bibinfo {booktitle} {Advances in Neural Information Processing
  Systems}}}\ (\bibinfo {year} {1992})\BibitemShut {NoStop}%
\bibitem [{\citenamefont {Wojtowytsch}(2021)}]{Wojtowytsch2021}%
  \BibitemOpen
  \bibfield  {author} {\bibinfo {author} {\bibfnamefont {Stephan}\ \bibnamefont
  {Wojtowytsch}},\ }\bibfield  {title} {\bibinfo {title} {{Stochastic gradient
  descent with noise of machine learning type. Part II: Continuous time
  analysis}},\ }\href {http://arxiv.org/abs/2106.02588} {\bibfield  {journal}
  {\bibinfo  {journal} {arXiv:2106.02588}\ } (\bibinfo {year}
  {2021})}\BibitemShut {NoStop}%
\bibitem [{\citenamefont {Pesme}\ \emph {et~al.}(2021)\citenamefont {Pesme},
  \citenamefont {Pillaud-Vivien},\ and\ \citenamefont
  {Flammarion}}]{Pesme2021}%
  \BibitemOpen
  \bibfield  {author} {\bibinfo {author} {\bibfnamefont {Scott}\ \bibnamefont
  {Pesme}}, \bibinfo {author} {\bibfnamefont {Loucas}\ \bibnamefont
  {Pillaud-Vivien}},\ and\ \bibinfo {author} {\bibfnamefont {Nicolas}\
  \bibnamefont {Flammarion}},\ }\bibfield  {title} {\bibinfo {title} {{Implicit
  Bias of SGD for Diagonal Linear Networks: a Provable Benefit of
  Stochasticity}},\ }\href {http://arxiv.org/abs/2106.09524} {\bibfield
  {journal} {\bibinfo  {journal} {arXiv:2106.09524}\ } (\bibinfo {year}
  {2021})}\BibitemShut {NoStop}%
\bibitem [{\citenamefont {G\"urb\"uzbalaban}\ \emph {et~al.}(2021)\citenamefont
  {G\"urb\"uzbalaban}, \citenamefont {^^c5^^9eim^^c5^^9fekli},\ and\
  \citenamefont {Zhu}}]{Gurbuzbalaban2021}%
  \BibitemOpen
  \bibfield  {author} {\bibinfo {author} {\bibfnamefont {Mert}\ \bibnamefont
  {G\"urb\"uzbalaban}}, \bibinfo {author} {\bibfnamefont {Umut}\ \bibnamefont
  {^^c5^^9eim^^c5^^9fekli}},\ and\ \bibinfo {author} {\bibfnamefont
  {Lingjiong}\ \bibnamefont {Zhu}},\ }\bibfield  {title} {\bibinfo {title}
  {{The Heavy-Tail Phenomenon in SGD}},\ }in\ \href
  {http://arxiv.org/abs/2006.04740} {\emph {\bibinfo {booktitle} {International
  Conference on Machine Learning}}}\ (\bibinfo {year} {2021})\BibitemShut
  {NoStop}%
\bibitem [{\citenamefont {Li}\ \emph {et~al.}(2017)\citenamefont {Li},
  \citenamefont {Tai},\ and\ \citenamefont {Weinan}}]{Li2017}%
  \BibitemOpen
  \bibfield  {author} {\bibinfo {author} {\bibfnamefont {Qianxiao}\
  \bibnamefont {Li}}, \bibinfo {author} {\bibfnamefont {Cheng}\ \bibnamefont
  {Tai}},\ and\ \bibinfo {author} {\bibfnamefont {E.}~\bibnamefont {Weinan}},\
  }\bibfield  {title} {\bibinfo {title} {{Stochastic modified equations and
  adaptive stochastic gradient algorithms}},\ }in\ \href
  {http://proceedings.mlr.press/v70/li17f.html} {\emph {\bibinfo {booktitle}
  {International Conference on Machine Learning}}}\ (\bibinfo {year}
  {2017})\BibitemShut {NoStop}%
\bibitem [{\citenamefont {Smith}\ and\ \citenamefont
  {Le}(2018)}]{Smith-Le2018}%
  \BibitemOpen
  \bibfield  {author} {\bibinfo {author} {\bibfnamefont {Samuel~L.}\
  \bibnamefont {Smith}}\ and\ \bibinfo {author} {\bibfnamefont {Quoc~V.}\
  \bibnamefont {Le}},\ }\bibfield  {title} {\bibinfo {title} {{A Bayesian
  perspective on generalization and stochastic gradient descent}},\ }in\ \href
  {https://openreview.net/forum?id=BJij4yg0Z} {\emph {\bibinfo {booktitle}
  {International Conference on Learning Representations}}}\ (\bibinfo {year}
  {2018})\BibitemShut {NoStop}%
\bibitem [{\citenamefont {Sato}\ and\ \citenamefont
  {Nakagawa}(2014)}]{Sato2014}%
  \BibitemOpen
  \bibfield  {author} {\bibinfo {author} {\bibfnamefont {Issei}\ \bibnamefont
  {Sato}}\ and\ \bibinfo {author} {\bibfnamefont {Hiroshi}\ \bibnamefont
  {Nakagawa}},\ }\bibfield  {title} {\bibinfo {title} {{Approximation analysis
  of stochastic gradient langevin dynamics by using fokker-planck equation and
  ito process}},\ }in\ \href {http://proceedings.mlr.press/v32/satoa14.html}
  {\emph {\bibinfo {booktitle} {International Conference on Machine
  Learning}}}\ (\bibinfo {year} {2014})\BibitemShut {NoStop}%
\bibitem [{\citenamefont {Zhang}\ \emph
  {et~al.}(2017{\natexlab{a}})\citenamefont {Zhang}, \citenamefont {Liang},\
  and\ \citenamefont {Charikar}}]{Zhang2017a}%
  \BibitemOpen
  \bibfield  {author} {\bibinfo {author} {\bibfnamefont {Yuchen}\ \bibnamefont
  {Zhang}}, \bibinfo {author} {\bibfnamefont {Percy}\ \bibnamefont {Liang}},\
  and\ \bibinfo {author} {\bibfnamefont {Moses}\ \bibnamefont {Charikar}},\
  }\bibfield  {title} {\bibinfo {title} {{A hitting time analysis of stochastic
  gradient Langevin dynamics}},\ }in\ \href@noop {} {\emph {\bibinfo
  {booktitle} {Proceedings of Machine Learning Research}}}\ (\bibinfo {year}
  {2017})\BibitemShut {NoStop}%
\bibitem [{\citenamefont {Ziyin}\ \emph {et~al.}(2021)\citenamefont {Ziyin},
  \citenamefont {Liu}, \citenamefont {Mori},\ and\ \citenamefont
  {Ueda}}]{Ziyin2021}%
  \BibitemOpen
  \bibfield  {author} {\bibinfo {author} {\bibfnamefont {Liu}\ \bibnamefont
  {Ziyin}}, \bibinfo {author} {\bibfnamefont {Kangqiao}\ \bibnamefont {Liu}},
  \bibinfo {author} {\bibfnamefont {Takashi}\ \bibnamefont {Mori}},\ and\
  \bibinfo {author} {\bibfnamefont {Masahito}\ \bibnamefont {Ueda}},\
  }\bibfield  {title} {\bibinfo {title} {{On Minibatch Noise: Discrete-Time
  SGD, Overparametrization, and Bayes}},\ }\href
  {https://arxiv.org/abs/2102.05375} {\bibfield  {journal} {\bibinfo  {journal}
  {arXiv:2102.05375}\ } (\bibinfo {year} {2021})}\BibitemShut {NoStop}%
\bibitem [{\citenamefont {Papyan}(2018)}]{Papyan2018}%
  \BibitemOpen
  \bibfield  {author} {\bibinfo {author} {\bibfnamefont {Vardan}\ \bibnamefont
  {Papyan}},\ }\bibfield  {title} {\bibinfo {title} {{The Full Spectrum of
  Deepnet Hessians at Scale: Dynamics with SGD Training and Sample Size}},\
  }\href {https://arxiv.org/abs/1811.07062} {\bibfield  {journal} {\bibinfo
  {journal} {arXiv:1811.07062}\ } (\bibinfo {year} {2018})}\BibitemShut
  {NoStop}%
\bibitem [{\citenamefont {{\O}ksendal}(1998)}]{Oksendal_text}%
  \BibitemOpen
  \bibfield  {author} {\bibinfo {author} {\bibfnamefont {Bernt}\ \bibnamefont
  {{\O}ksendal}},\ }\href@noop {} {\emph {\bibinfo {title} {{Stochastic
  differential equations: an introduction with applications}}}}\ (\bibinfo
  {publisher} {Springer},\ \bibinfo {address} {Berlin},\ \bibinfo {year}
  {1998})\BibitemShut {NoStop}%
\bibitem [{\citenamefont {Zhang}\ \emph
  {et~al.}(2017{\natexlab{b}})\citenamefont {Zhang}, \citenamefont {Bengio},
  \citenamefont {Hardt}, \citenamefont {Recht},\ and\ \citenamefont
  {Vinyals}}]{Zhang2017}%
  \BibitemOpen
  \bibfield  {author} {\bibinfo {author} {\bibfnamefont {Chiyuan}\ \bibnamefont
  {Zhang}}, \bibinfo {author} {\bibfnamefont {Samy}\ \bibnamefont {Bengio}},
  \bibinfo {author} {\bibfnamefont {Moritz}\ \bibnamefont {Hardt}}, \bibinfo
  {author} {\bibfnamefont {Benjamin}\ \bibnamefont {Recht}},\ and\ \bibinfo
  {author} {\bibfnamefont {Oriol}\ \bibnamefont {Vinyals}},\ }\bibfield
  {title} {\bibinfo {title} {{Understanding Deep Learning Requires Rethinking
  of Generalization}},\ }in\ \href
  {https://openreview.net/forum?id=Sy8gdB9xx&;amp;noteId=Sy8gdB9xx} {\emph
  {\bibinfo {booktitle} {International Conference on Learning
  Representations}}}\ (\bibinfo {year} {2017})\BibitemShut {NoStop}%
\bibitem [{\citenamefont {Langer}(1969)}]{Langer1969}%
  \BibitemOpen
  \bibfield  {author} {\bibinfo {author} {\bibfnamefont {James~S.}\
  \bibnamefont {Langer}},\ }\bibfield  {title} {\bibinfo {title} {{Statistical
  theory of the decay of metastable states}},\ }\href
  {https://doi.org/10.1016/0003-4916(69)90153-5} {\bibfield  {journal}
  {\bibinfo  {journal} {Annals of Physics}\ }\textbf {\bibinfo {volume} {54}},\
  \bibinfo {pages} {258--275} (\bibinfo {year} {1969})}\BibitemShut {NoStop}%
\bibitem [{\citenamefont {Bovier}\ \emph {et~al.}(2004)\citenamefont {Bovier},
  \citenamefont {Eckhoff}, \citenamefont {Gayrard},\ and\ \citenamefont
  {Klein}}]{Bovier2004}%
  \BibitemOpen
  \bibfield  {author} {\bibinfo {author} {\bibfnamefont {Anton}\ \bibnamefont
  {Bovier}}, \bibinfo {author} {\bibfnamefont {Michael}\ \bibnamefont
  {Eckhoff}}, \bibinfo {author} {\bibfnamefont {V{\'{e}}ronique}\ \bibnamefont
  {Gayrard}},\ and\ \bibinfo {author} {\bibfnamefont {Markus}\ \bibnamefont
  {Klein}},\ }\bibfield  {title} {\bibinfo {title} {{Metastability in
  reversible diffusion processes I. Sharp asymptotics for capacities and exit
  times}},\ }\href {https://doi.org/10.4171/JEMS/14} {\bibfield  {journal}
  {\bibinfo  {journal} {Journal of the European Mathematical Society}\ }\textbf
  {\bibinfo {volume} {6}},\ \bibinfo {pages} {399--424} (\bibinfo {year}
  {2004})}\BibitemShut {NoStop}%
\bibitem [{\citenamefont {Berglund}(2013)}]{Berglund2013}%
  \BibitemOpen
  \bibfield  {author} {\bibinfo {author} {\bibfnamefont {Nils}\ \bibnamefont
  {Berglund}},\ }\bibfield  {title} {\bibinfo {title} {{Kramers' Law: Validity,
  Derivations and Generalisations}},\ }\href {https://arxiv.org/abs/1106.5799}
  {\bibfield  {journal} {\bibinfo  {journal} {Markov Processes and Related
  Fields}\ }\textbf {\bibinfo {volume} {19}},\ \bibinfo {pages} {459--490}
  (\bibinfo {year} {2013})}\BibitemShut {NoStop}%
\bibitem [{\citenamefont {Bianchi}\ and\ \citenamefont
  {Gaudilli{\`{e}}re}(2016)}]{Bianchi2016}%
  \BibitemOpen
  \bibfield  {author} {\bibinfo {author} {\bibfnamefont {Alessandra}\
  \bibnamefont {Bianchi}}\ and\ \bibinfo {author} {\bibfnamefont {Alexandre}\
  \bibnamefont {Gaudilli{\`{e}}re}},\ }\bibfield  {title} {\bibinfo {title}
  {{Metastable states, quasi-stationary distributions and soft measures}},\
  }\href {https://doi.org/10.1016/j.spa.2015.11.015} {\bibfield  {journal}
  {\bibinfo  {journal} {Stochastic Processes and their Applications}\ }\textbf
  {\bibinfo {volume} {126}},\ \bibinfo {pages} {1622--1680} (\bibinfo {year}
  {2016})}\BibitemShut {NoStop}%
\bibitem [{\citenamefont {Freidlin}\ and\ \citenamefont
  {Wentzell}(1998)}]{Freidlin_text}%
  \BibitemOpen
  \bibfield  {author} {\bibinfo {author} {\bibfnamefont {Mark~I.}\ \bibnamefont
  {Freidlin}}\ and\ \bibinfo {author} {\bibfnamefont {Alexander~D.}\
  \bibnamefont {Wentzell}},\ }\href@noop {} {\emph {\bibinfo {title} {{Random
  Perturbations of Dynamical Systems}}}}\ (\bibinfo  {publisher} {Springer},\
  \bibinfo {year} {1998})\BibitemShut {NoStop}%
\bibitem [{\citenamefont {Maier}\ and\ \citenamefont
  {Stein}(1993)}]{Maier1993}%
  \BibitemOpen
  \bibfield  {author} {\bibinfo {author} {\bibfnamefont {Robert~S}\
  \bibnamefont {Maier}}\ and\ \bibinfo {author} {\bibfnamefont {Daniel~L.}\
  \bibnamefont {Stein}},\ }\bibfield  {title} {\bibinfo {title} {{Escape
  problem for irreversible systems}},\ }\href
  {https://doi.org/10.1103/PhysRevE.48.931} {\bibfield  {journal} {\bibinfo
  {journal} {Physical Review E}\ }\textbf {\bibinfo {volume} {48}},\ \bibinfo
  {pages} {931--938} (\bibinfo {year} {1993})}\BibitemShut {NoStop}%
\bibitem [{\citenamefont {Gur-Ari}\ \emph {et~al.}(2018)\citenamefont
  {Gur-Ari}, \citenamefont {Roberts},\ and\ \citenamefont
  {Dyer}}]{Gur-Ari2018}%
  \BibitemOpen
  \bibfield  {author} {\bibinfo {author} {\bibfnamefont {Guy}\ \bibnamefont
  {Gur-Ari}}, \bibinfo {author} {\bibfnamefont {Daniel~A.}\ \bibnamefont
  {Roberts}},\ and\ \bibinfo {author} {\bibfnamefont {Ethan}\ \bibnamefont
  {Dyer}},\ }\bibfield  {title} {\bibinfo {title} {{Gradient Descent Happens in
  a Tiny Subspace}},\ }\href {http://arxiv.org/abs/1812.04754} {\bibfield
  {journal} {\bibinfo  {journal} {arXiv:1812.04754}\ } (\bibinfo {year}
  {2018})}\BibitemShut {NoStop}%
\bibitem [{\citenamefont {Li}\ \emph {et~al.}(2018)\citenamefont {Li},
  \citenamefont {Farkhoor}, \citenamefont {Liu},\ and\ \citenamefont
  {Yosinski}}]{Li2018}%
  \BibitemOpen
  \bibfield  {author} {\bibinfo {author} {\bibfnamefont {Chunyuan}\
  \bibnamefont {Li}}, \bibinfo {author} {\bibfnamefont {Heerad}\ \bibnamefont
  {Farkhoor}}, \bibinfo {author} {\bibfnamefont {Rosanne}\ \bibnamefont
  {Liu}},\ and\ \bibinfo {author} {\bibfnamefont {Jason}\ \bibnamefont
  {Yosinski}},\ }\bibfield  {title} {\bibinfo {title} {{Measuring the Intrinsic
  Dimension of Objective Landscapes}},\ }in\ \href
  {https://openreview.net/forum?id=ryup8-WCW} {\emph {\bibinfo {booktitle}
  {International Conference on Learning Representations}}}\ (\bibinfo {year}
  {2018})\BibitemShut {NoStop}%
\bibitem [{\citenamefont {^^c5^^9eim^^c5^^9fekli}\ \emph
  {et~al.}(2019)\citenamefont {^^c5^^9eim^^c5^^9fekli}, \citenamefont {Sagun},\
  and\ \citenamefont {Giirbiizbalaban}}]{Simsekli2019}%
  \BibitemOpen
  \bibfield  {author} {\bibinfo {author} {\bibfnamefont {Umut}\ \bibnamefont
  {^^c5^^9eim^^c5^^9fekli}}, \bibinfo {author} {\bibfnamefont {Levent}\
  \bibnamefont {Sagun}},\ and\ \bibinfo {author} {\bibfnamefont {Mert}\
  \bibnamefont {Giirbiizbalaban}},\ }\bibfield  {title} {\bibinfo {title} {{A
  tail-index analysis of stochastic gradient noise in deep neural networks}},\
  }in\ \href {http://proceedings.mlr.press/v97/simsekli19a.html} {\emph
  {\bibinfo {booktitle} {International Conference on Machine Learning}}}\
  (\bibinfo {year} {2019})\BibitemShut {NoStop}%
\end{thebibliography}%

\newpage
\appendix
\onecolumn
\section{List of approximations and their justifications}
\label{appendix:list}

In \cref{sec:SGD_noise}, we made several approximations to derive the result~(\ref{eq:SGD_cov_formula}).
For clarity, we list the approximations made and their justifications below.

\begin{itemize}
\item In Eq.~(\ref{eq:SGD_cov}), we make the approximation of $B\ll N$, which simplifies the expression but is not essential. The main conclusion is not affected by this approximation.
\item We ignore $\nabla L\nabla L^\mathrm{T}$ in Eq.~(\ref{eq:SGD_cov}), which is justified near a local minimum.
\item In Eq.~(\ref{eq:SGD_cov_MSE}), we make the decoupling approximation~(\ref{eq:decoupling}), which is one of the key heuristic approximations in our work. This approximation is verified experimentally in \cref{sec:decoupling}.
\item We ignore the last term of Eq.~(\ref{eq:Hessian}), which is a common approximation~\citep{Sagun2017}. This approximation is justified if we are only interested in the outliers of the Hessian eigenvalues. 
Indeed, outliers of the Hessian eigenvalues, which play dominant roles in escape from local minima, are known to be attributable to the first term of the right-hand side of Eq.~(\ref{eq:Hessian})~\citep{Papyan2018}.
\item In Eq.~(\ref{eq:Hessian_approx}), we assume that the matrix $(1/N)\sum_{\mu=1}^N\nabla f(\theta,x^{(\mu)})\nabla f(\theta,x^{(\mu)})^T$ does not change so much in a valley with a given local minimum.
This approximation is indirectly verified in Fig.~\ref{fig:noise} (the proportionality between the loss and the noise strength implies this matrix is actually constant). 
\end{itemize}

\section{Proof of \cref{theo:main}}
\label{sec:proof}

We now prove \cref{theo:main}.
Under \cref{theo:subspace}, dynamics of $\theta$ is restricted to the $n$-dimensional subspace spanned by the outlier eigenvectors $v_1,v_2,\dots,v_n\in\mathbb{R}^P$ of the Hessian.
Consequently, $\theta$ is parametrized by $n$ variables $z_1,z_2,\dots,z_n\in\mathbb{R}$ as
\begin{equation}
\theta=\theta^*+\sum_{i=1}^nz_iv_i.
\end{equation}
The Hessian is a diagonal matrix in the basis of $(v_1,v_2,\dots,v_n)$ within $\mathcal{A}_{\theta^*}$ because of \cref{theo:subspace}. 
Therefore, $\del^2U/\del z_i\del z_j=0$ for any $i\neq j$, where $U=\log L$ is the logarithmic loss introduced in Eq.~(\ref{eq:SDE_tau}).
It implies that $\del U/\del z_i$ is a function of $z_i$, and hence $U$ is written in the form
\begin{equation}
U(\theta)=\log L(\theta^*)+\sum_{i=1}^nU_i(z_i).
\end{equation}
It should be noted that
\begin{equation}
U_i(0)=0, \quad U_i(z_i)>0 \text{ for any $z_i\neq 0$},\quad \frac{\del U_i}{\del z_i}(0)=0, \quad \frac{\del^2U_i}{\del z_i^2}(0)=\frac{h_i^*}{L(\theta^*)},
\label{eq:U_i}
\end{equation}
where the last equality is obtained by putting $U=\log L$ and $\left.\del^2 L/\del z_i^2\right|_{z=0}=h_i^*$.

The stochastic differential equation for the $n$-dimensional vector $z=(z_1,z_2,\dots,z_n)^\mathrm{T}$ in the rescaled time variable $\tau$ is given by $dz=-\nabla_z Ud\tau+\sqrt{2\eta \hat{H}(\theta^*)/B}d\tilde{W}_\tau$ [see Eq.~(\ref{eq:SDE_tau}), which is equivalent to the following Fokker-Planck equation for the distribution function $P(z,\tau)$ of $z$ at $\tau$:
\begin{equation}
\frac{\del P(z,\tau)}{\del\tau}=\sum_{i=1}^n\left[\frac{\del}{\del z_i}\left(\frac{\del U_i}{\del z_i}P\right)+\frac{\eta h_i^*}{B}\frac{\del^2}{\del z_i^2}P\right]=:-\nabla_z J(z),
\label{eq:FP_tau}
\end{equation}
where the probability current density $J(z)=(J_1(z),J_2(z),\dots,J_n(z))^\mathrm{T}$ is explicitly given by
\begin{equation}
J_i(z)=-\frac{\del U_i(z_i)}{\del z_i}P(z)-\frac{\eta h_i^*}{B}\frac{\del}{\del z_i}P(z).
\end{equation}

Under \cref{theo:MPEP}, the escape rate is evaluated by considering the MPEP.
It is known that the MPEP aligns with one of the eigenvectors of the Hessian~\citep{Xie2021}.
Without loss of generality, let us assume that the direction of $e$th eigenvector $v_e$ corresponds to the direction of the MPEP.
We denote by $z_\perp\in\mathbb{R}^{n-1}$ the displacement perpendicular to the escape direction, and write $z=(z_e,z_\perp)$.
At the saddle $z^s$, $h_e^s<0$ and $h_i^s>0$ for all $i\neq e$, where $\{h_i^s\}$ is the set of eigenvalues of the Hessian at $z^s$, and
\begin{equation}
\frac{\del U_i}{\del z_i}(z_i^s)=0, \quad \frac{\del^2 U_i}{\del z_i^2}(z_i^s)=\frac{h_i^s}{L(\theta^s)}.
\end{equation}

We now derive the escape rate formula for asymptotically weak noise $\eta/B\to +0$ following \citet{Kramers1940}.
The steady current $J\in\mathbb{R}^n$ is aligned to the escape direction, and hence $J_e\neq 0$ and $J_\perp=0$ along the MPEP.
Let us denote by $P^*$ the total probability within the valley $\mathcal{A}_{\theta^*}$ and by $\mathcal{J}_\tau$ the total current per unit change of $\tau$ flowing to the outside of the valley through the saddle $\theta^s=\theta^*+z^s$.
From \cref{theo:qss}, $z$ is assumed to be in the quasi-stationary distribution.
Since the quasi-stationary distribution is concentrated around $\theta^*$ in the weak-noise limit, $L(\theta_t)$ is approximately equal to $L(\theta^*)$ before the escape.
Consequently, the rescaled time variable $\tau$ in Eq.~(\ref{eq:tau}) is approximately equal to $L(\theta^*)t$, and hence the total probability current $\mathcal{J}_t$ per unit change of $t$ is given by $L(\theta^*)\mathcal{J}_\tau$.
The escape rate $\kappa$ per unit change of $t$ is therefore given by
\begin{equation}
\kappa=\frac{\mathcal{J}_t}{P^*}\sim L(\theta^*)\frac{\mathcal{J}_\tau}{P^*} \quad\text{as }\frac{\eta}{B}\to +0.\footnote{Recall that the notation ``$f(x)\sim g(x)$ as $x\to +0$'' means $\lim_{x\to +0}(f(x)/g(x))=1$.}
\label{eq:kappa}
\end{equation}

We now evaluate $P^*$ and $\mathcal{J}_\tau$.
First, to obtain $P^*$, we must know about the quasi-stationary distribution $P_\mathrm{s}(z)$ near $\theta^*$.
It is obtained by putting $\del P(z,\tau)/\del\tau=0$ in Eq.~(\ref{eq:FP_tau}), which implies $J(z)=0$.
We therefore have
\begin{equation}
\frac{\del P_\mathrm{s}(z)}{\del z_i}=-\frac{B}{\eta h_i^*}\frac{\del U_i(z_i)}{\del z_i}P_\mathrm{s}(z).
\end{equation}
By solving this equation, we obtain for $\theta=\theta^*+z\in\mathcal{A}_{\theta^*}$
\begin{equation}
P_\mathrm{s}(z)=P(\theta^*)\exp\left[-\sum_{i=1}^n\frac{B}{\eta h_i^*}U_i(z_i)\right].
\label{eq:P_ss}
\end{equation}
For convenience, we put $P_\mathrm{s}(z)=0$ for $z\notin\mathcal{A}_{\theta^*}$.
Then, $P^*$, which is the total probability within $\mathcal{A}_{\theta^*}$, is given by
\begin{equation}
P^*=\int_{-\infty}^\infty dz_1\int_{-\infty}^\infty dz_2\dots\int_{-\infty}^\infty dz_n\, P_\mathrm{s}(z)
=P(\theta^*)\prod_{i=1}^n\int_{-\infty}^\infty dz_i\, e^{-\frac{B}{\eta h_i^*}U(z_i)}.
\end{equation}
In the weak-noise limit $\eta/B\to +0$, we can evaluate the above integral by using the saddle-point method, which is also called the Laplace method.
By using Eq.~(\ref{eq:U_i}), the saddle-point method yields
\begin{equation}
\int_{-\infty}^\infty dz_i\, e^{-\frac{B}{\eta h_i^*}U(z_i)}\sim\int_{-\infty}^\infty dz_i\, e^{-\frac{B}{2\eta L(\theta^*)}z_i^2}=\left[\frac{2\pi\eta L(\theta^*)}{B}\right]^{1/2}
\end{equation}
as $\eta/B\to +0$, and therefore
\begin{equation}
P^*\sim P(\theta^*)\left[\frac{2\pi\eta L(\theta^*)}{B}\right]^{n/2}.
\label{eq:P_final}
\end{equation}

\begin{figure}[tb]
\centering
\includegraphics[width=0.7\linewidth]{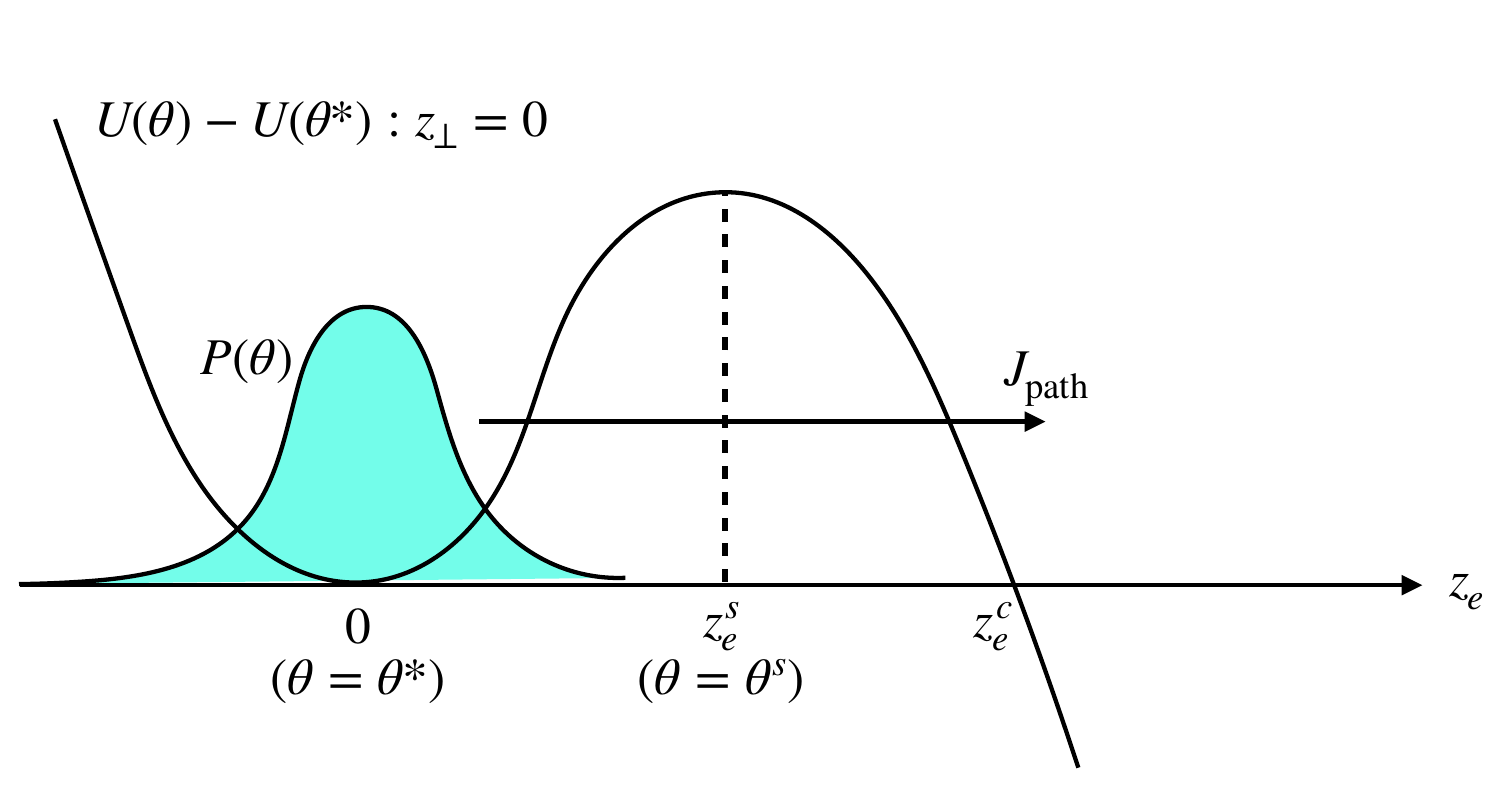}
\caption{Schematic illustration of the escape from a potential barrier.}
\label{fig:potential}
\end{figure}

Next, let us evaluate $\mathcal{J}_\tau$.
Along the MPEP ($z_\perp=0$), the current vector $J_\perp$ perpendicular to the escape direction is zero, and hence the probability distribution near the saddle is evaluated in a similar way as Eq.~(\ref{eq:P_ss}):
\begin{equation}
P(z_e=z_e^s,z_\perp)=P(z_e=z_e^s,z_\perp=0)\exp\left[-\sum_{i(\neq e)}\frac{B}{\eta h_i^*}U_i(z_i)\right].
\end{equation}
By substituting it into
\begin{equation}
J_e=-\frac{\del U_e}{\del z_e}P-\frac{\eta h_e^*}{B}\frac{\del}{\del z_e}P,
\end{equation}
we obtain
\begin{equation}
J_e(z_e^s,z_\perp)=J_\mathrm{path}\exp\left[-\sum_{i(\neq e)}\frac{B}{\eta h_i^*}U_i(z_i)\right],
\label{eq:J_e}
\end{equation}
where the current along the MPEP ($z_\perp=0$) is denoted by $J_\mathrm{path}:=J(z_e^s,z_\perp=0)$.
The total current through the saddle is then evaluated by using the saddle-point method:
\begin{equation}
\mathcal{J}_\tau=\int dz_\perp\, J_e(z_e^s,z_\perp)=J_\mathrm{path}\prod_{i(\neq e)}\int_{-\infty}^\infty dz_\perp\, e^{-\frac{B}{\eta h_i^*}U_i(z_i)}\sim J_\mathrm{path}\left[\frac{2\pi\eta L(\theta^s)}{B}\right]^{(n-1)/2}
\label{eq:total_J}
\end{equation}
as $\eta/B\to +0$.
When the distribution function is almost stationary, Eq.~(\ref{eq:FP_tau}) yields $\del J_e(z_e,z_\perp=0)/\del z_e=0$, and hence the current along the MPEP is constant $J_e(z_e,z_\perp=0)=J_\mathrm{path}$.
By putting $z_\perp=0$ in Eq.~(\ref{eq:J_e}), we have
\begin{equation}
J_\mathrm{path}=-\frac{\del U}{\del z_e}P(z_e,z_\perp=0)-\frac{\eta h_e^*}{B}\frac{\del P}{\del z_e}(z_e,z_\perp=0)
=-\frac{\eta h_e^*}{B}e^{-\frac{B}{\eta h_e^*}U}\frac{\del}{\del z_e}\left[e^{\frac{B}{\eta h_e^*}U}P(z_e,z_\perp=0)\right].
\end{equation}
By multiplying $e^{\frac{B}{\eta h_e^*}U}$ in both sides and integrating over $z_e$ from 0 to $z_e^c$, where $z_e^c$ defined as $U_e(\theta^*+z_e^cv_e)=U_e(\theta^*)$ (see Fig.~\ref{fig:potential}), we obtain
\begin{equation}
J_\mathrm{path}\int_0^{z_e^c}dz_e\, e^{\frac{B}{\eta h_e^*}U}\sim\frac{\eta h_e^*}{B}e^{\frac{B}{\eta h_e^*}U(\theta^*)}P(\theta^*)
\label{eq:J_int}
\end{equation}
as $\eta/B\to +0$, where we used the fact that the probability at $z_e^c$ is negligible in the weak-noise limit.
By using the saddle-point method, the integral in the left-hand side of Eq.~(\ref{eq:J_int}) is evaluated as
\begin{align}
\int_0^{z_e^c}dz_e\, e^{\frac{B}{\eta h_e^*}U}&\sim\int_{-\infty}^\infty dz_e\, \exp\left[\frac{B}{\eta h_e^*}\left(U(\theta^s)+\frac{h_e^s}{2L(\theta^s)}(z_e-z_e^s)^2\right)\right]
\nonumber \\
&=\left(\frac{2\pi\eta h_e^*}{B|h_e^s|L(\theta^s)}\right)^{1/2}e^{\frac{B}{\eta h_e^*}U(\theta^s)}.
\end{align}
By substituting this result in Eq.~(\ref{eq:J_int}), we obtain
\begin{equation}
J_\mathrm{path}\sim\left(\frac{\eta h_e^*|h_e^s|}{2\pi BL(\theta^s)}\right)^{1/2}e^{-\frac{B}{\eta h_e^*}\Delta U}P(\theta^*),
\end{equation}
where $\Delta U=U(\theta^s)-U(\theta^*)$.
The total current $\mathcal{J}_\tau$ in Eq.~(\ref{eq:total_J}) is then expressed as
\begin{equation}
\mathcal{J}_\tau\sim\frac{\sqrt{h_e^*|h_e^s|}}{2\pi L(\theta^s)}\left(\frac{2\pi\eta L(\theta^s)}{B}\right)^{n/2}e^{-\frac{B}{\eta h_e^*}\Delta U}P(\theta^*).
\label{eq:J_total_final}
\end{equation}
By using Eqs.~(\ref{eq:P_final}) and (\ref{eq:J_total_final}), the escape rate $\kappa$ in Eq.~(\ref{eq:kappa}) is evaluated as
\begin{equation}
\kappa\sim L(\theta^*)\frac{\mathcal{J}_\tau}{P^*}=\frac{\sqrt{h_e^*|h_e^s|}}{2\pi}\left[\frac{L(\theta^s)}{L(\theta^*)}\right]^{\frac{n}{2}-1}e^{-\frac{B}{\eta h_e*}\Delta U}.
\end{equation}
Since $\Delta U=\log[L(\theta^s)/L(\theta^*)]$, we finally obtain
\begin{equation}
\kappa\sim\frac{\sqrt{h_e^*|h_e^s|}}{2\pi}\left[\frac{L(\theta^s)}{L(\theta^*)}\right]^{-\left(\frac{B}{\eta h_e^*}+1-\frac{n}{2}\right)},
\end{equation}
which is exactly identical to the escape rate formula given in \cref{theo:main}.

\section{Stationary distribution}
\label{appendix:stationary}

Suppose that $\tilde{\theta}_\tau$ obeys a simple Langevin equation
\begin{equation}
d\tilde{\theta}_\tau=-U'(\tilde{\theta}_\tau)+\sqrt{2T}d\tilde{W}_\tau.
\end{equation}
The stationary distribution of $\tilde{\theta}_\tau$ is then given by the Gibbs distribution $\tilde{P}_\mathrm{s}(\theta)\propto e^{-U(\theta)/T}$.
On the other hand, what we want is the stationary distribution $P_\mathrm{s}(\theta)$ of $\theta_t$, where $\theta_t=\tilde{\theta}_\tau$ with $\tau=\int_0^tdt'\, L(\theta_{t'})$.
In this section, we show the relation between the two distributions: $P_\mathrm{s}(\theta)\propto L(\theta)^{-1}\tilde{P}_\mathrm{s}(\theta)$.

We express the stationary distributions in terms of the long-time average of the delta function:
\begin{equation}
P_\mathrm{s}(\theta)=\lim_{s\to\infty}\frac{1}{s}\int_0^s dt\, \delta(\theta_t-\theta),
\quad \tilde{P}_\mathrm{s}(\theta)=\lim_{s\to\infty}\frac{1}{s}\int_0^s d\tau\, \delta(\tilde{\theta}_{\tau}-\theta).
\label{eq:long_time}
\end{equation}
By using the relation $\tau=\int_0^t dt'\, L(\theta_{t'})$, we have $d\tau=L(\theta_t)dt$.
For a sufficiently large $t$, we also obtain $\tau\sim t\bar{L}$, where $\bar{L}:=\lim_{s\to\infty}(1/s)\int_0^sdt'\,L(\theta_{t'})$ denotes the long-time average of $L(\theta_t)$.
By using them, $P_\mathrm{s}(\theta)$ is rewritten as
\begin{align}
P_\mathrm{s}(\theta)&\approx\lim_{s\to\infty}\frac{1}{s}\int_0^{s\bar{L}}d\tau\frac{\delta(\tilde{\theta}_\tau-\theta)}{L(\tilde{\theta}_\tau)}
\nonumber \\
&=\frac{1}{L(\theta)}\lim_{s\to\infty}\frac{\bar{L}}{s\bar{L}}\int_0^{s\bar{L}}d\tau\,\delta(\tilde{\theta}_\tau-\theta)
\nonumber \\
&=\frac{\bar{L}}{L(\theta)}\tilde{P}_\mathrm{s}(\theta).
\end{align}
We thus obtain the desired relation, $P_\mathrm{s}(\theta)\propto L(\theta)^{-1}\tilde{P}_\mathrm{s}(\theta)$.

\section{Other loss functions}
\label{appendix:loss}

\begin{figure}[t]
\centering
\begin{tabular}{cc}
(a) & (b) \\
\includegraphics[width=0.4\linewidth]{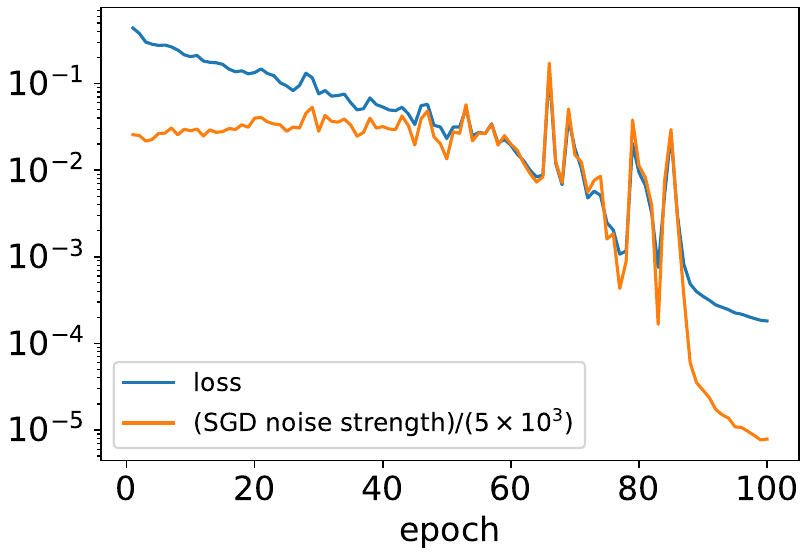}&
\includegraphics[width=0.4\linewidth]{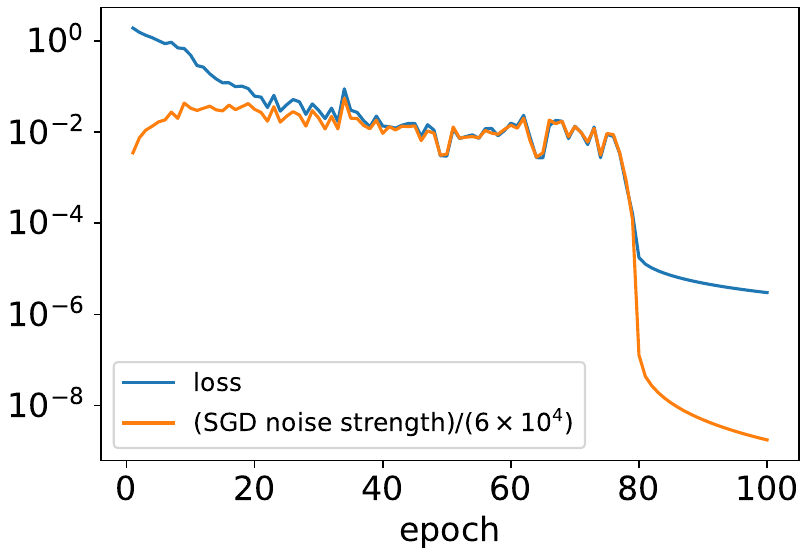}\\
(c) & (d) \\
\includegraphics[width=0.4\linewidth]{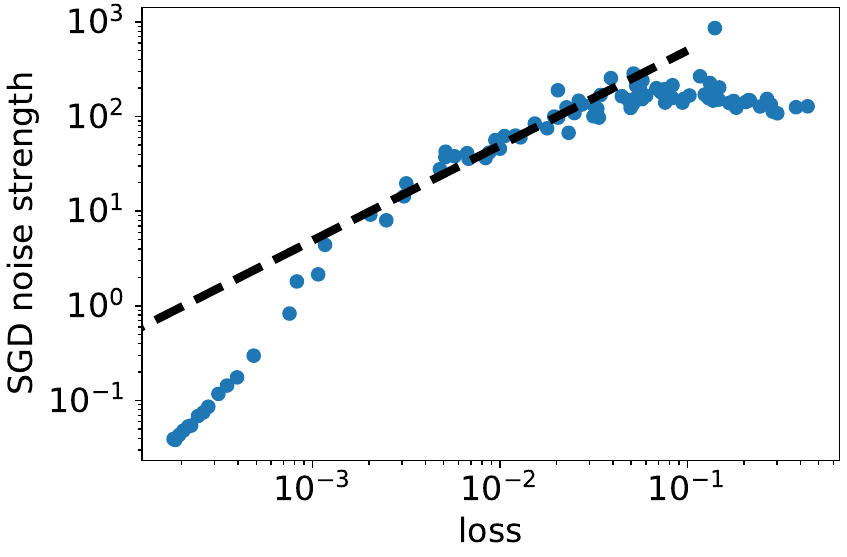}&
\includegraphics[width=0.4\linewidth]{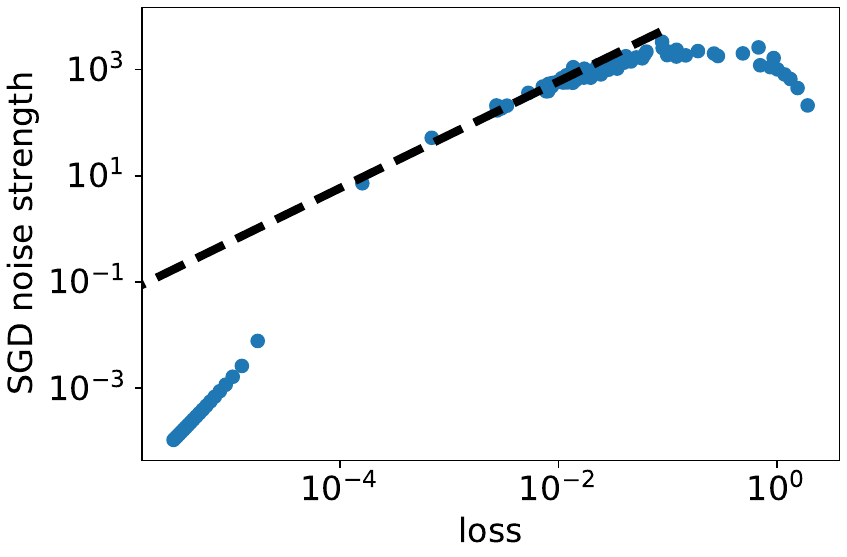}
\end{tabular}
\caption{Training dynamics of the cross-entropy loss and the SGD noise strength $\mathcal{N}$ for (a) a fully connected network trained by the Fashion-MNIST dataset and (b) a convolutional network trained by the CIFAR-10 dataset.
In the figure, we multiplied $\mathcal{N}$ by a numerical factor to emphasize that $\mathcal{N}$ is actually proportional to the loss in an intermediate stage of the training.
Loss vs $\mathcal{N}$ for (c) a fully connected network trained by the Fashion-MNIST and (d) a convolutional network trained by CIFAR-10.
Dashed lines in (c) and (d) are straight lines of slope 1, which imply $\mathcal{N}\propto L(\theta)$.}
\label{fig:other_loss}
\end{figure}

In our paper, we mainly focus on the mean-square loss, for which we can analytically derive the relation between the loss $L(\theta)$ and the SGD noise covariance $\Sigma(\theta)$.
An important observation is that the SGD noise strength $\mathcal{N}$ is proportional to the loss, i.e., $\mathcal{N}\propto L(\theta)$ (see \cref{sec:noise_strength} for the definition of $\mathcal{N}$).

Here, we argue that the relation $\mathcal{N}\propto L(\theta)$ also holds in more general situations.
During the training, the value of $\ell_\mu$ will fluctuate from sample to sample.
At a certain time step of SGD, let us suppose that $N-M$ samples in the training dataset are already fit correctly and hence $\ell_\mu\approx 0$, whereas the other $M$ samples are not and hence $\ell_\mu=\mathcal{O}(1)$.
The loss function is then given by $L(\theta)=(1/N)\sum_{\mu=1}^N\ell_\mu\propto M/N$.
When $\ell_\mu$ is small, $\nabla\ell_\mu$ will also be small.
Therefore, for $N-M$ samples with $\ell_\mu\approx 0$, $\nabla\ell_\mu\approx 0$ also holds.
The other $M$ samples will have non-small gradients: $\|\nabla\ell_\mu\|^2=\mathcal{O}(1)$.
We thus estimate $\mathcal{N}$ as $\mathcal{N}\approx(1/N)\sum_{\mu=1}^N\nabla\ell_\mu^\mathrm{T}\nabla\ell_\mu\propto M/N\propto L(\theta)$.
In this way, $\mathcal{N}\propto L(\theta)$ will hold, irrespective of the loss function.

However, we emphasize that this is a crude argument.
In particular, the above argument will not hold near a global minimum because all the samples are correctly fit there, which implies that $\ell_\mu$ is small for all $\mu$, in contrast to the above argument relying on the existence of $M$ samples with $\ell_\mu$ and $\|\nabla\ell_\mu\|^2$ of $\mathcal{O}(1)$.

We now experimentally test the relation $\mathcal{N}\propto L(\theta)$ for the cross-entropy loss.
We consider the same architectures and datasets in \cref{sec:noise_strength}: a fully connected neural network trained by Fashion-MNIST and a convolutional neural network trained by CIFAR-10 (see \cref{sec:noise_strength} for the detail).
We fix $B=100$ in both cases, and $\eta=0.1$ for the fully connected network and $\eta=0.05$ for the convolutional network.
Experimental results are presented in Fig.~\ref{fig:other_loss}.
We find that the relation $\mathcal{N}\propto L(\theta)$ seems to hold true at an intermediate stage of the training dynamics, although the proportionality is less clear compared with Fig.~\ref{fig:noise} in the main text for the mean-square loss.

We also find that for sufficiently small values of the loss, $\mathcal{N}\propto L(\theta)^2$ [see Fig.~\ref{fig:other_loss} (c) and (d)], whose implications should merit further investigation in future studies.

\section{Hessian eigenvalues for a neural network in \cref{sec:stationary}}
\label{sec:Hessian}

\begin{figure}[tb]
\centering
\begin{tabular}{cc}
(a) & (b) \\
\includegraphics[width=0.48\linewidth]{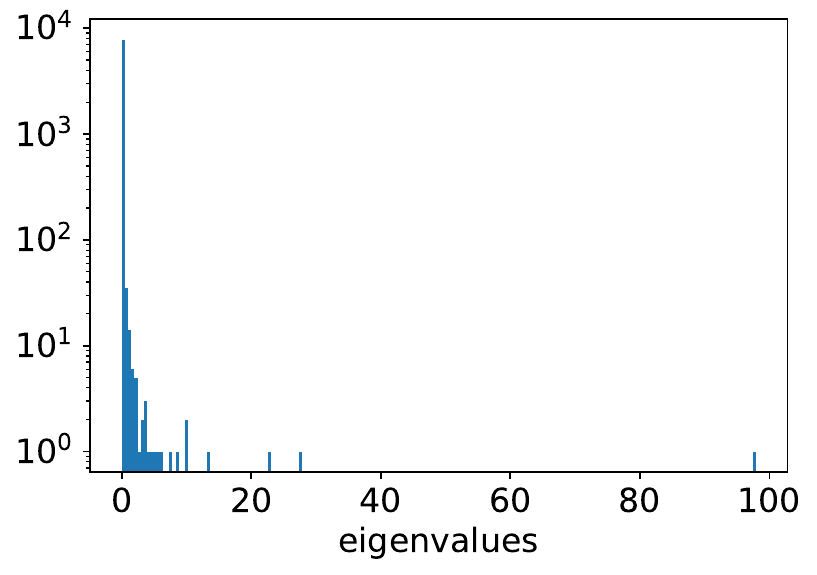}&
\includegraphics[width=0.48\linewidth]{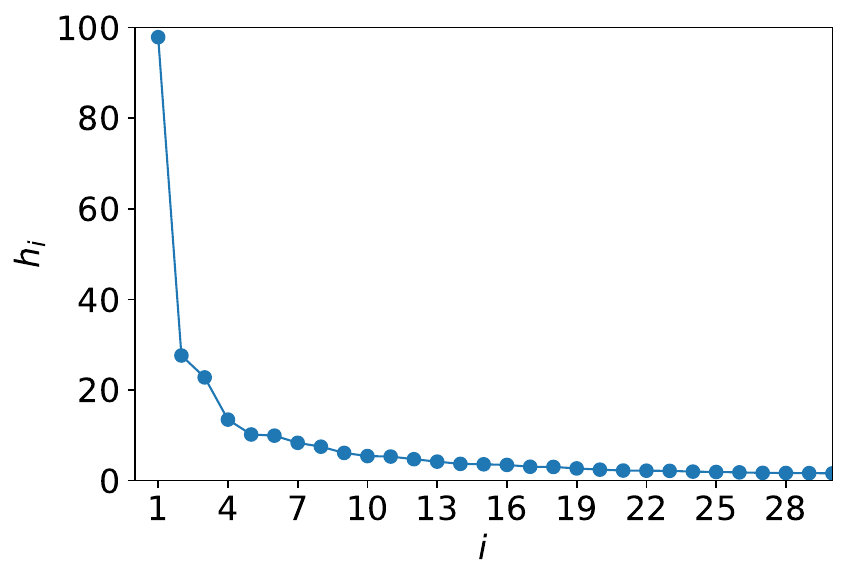}
\end{tabular}
\caption{Hessian eigenvalues for a pre-trained neural network studied in \cref{sec:stationary}.
(a) The histogram of the Hessian eigenvalues. Most eigenvalues are close to zero, but there are some large eigenvalues, which correspond to outliers.}
\label{fig:Hessian}
\end{figure}

We present numerical results on Hessian eigenvalues in a pre-trained neural network studied in \cref{sec:stationary}.
Instead of the exact Hessian, we consider an approximate Hessian given on the right-hand side of Eq.~(\ref{eq:Hessian_approx}), i.e.,
\begin{equation}
H(\theta^*)\approx\frac{1}{N}\sum_{\mu=1}^N\nabla f(\theta,x^{(\mu)})\nabla f(\theta,x^{(\mu)})^\mathrm{T}.
\end{equation}
Eigenvalues $\{h_i\}$ are arranged in descending order as $h_1\geq h_2\geq\dots\geq h_P$ (in our model $P=7861$).

A histogram of the Hessian eigenvalues is presented in Fig.~\ref{fig:Hessian} (a).
We see that most eigenvalues are close to zero, which corresponds to the bulk, but there are some large eigenvalues, which correspond to the outliers.
The largest eigenvalue is $\lambda_1=95.6$, which is identified as $h^*$ in our theoretical formula~(\ref{eq:escape_multi}).

Another important quantity is the effective dimension $n$ corresnding to the number of outliers.
Since the outliers and the bulk are not sharply separated, it is difficult to precisely determine $n$.
In Fig.~\ref{fig:Hessian} (b), we plot $h_i$ up to $i=30$.
From this figure, it seems reasonable to estimate $n\approx 10$.

As a heuristic method of determining $n$, we can consider the following identification: first we define the weight $p_i$ for $i$th eigenmode as $p_i=h_i/\sum_{j=1}^Ph_j$.
We then determine $n$ as
\begin{equation}
n=\left(\sum_{i=1}^Pp_i^2\right)^{-1},
\end{equation}
which gives $n=12.7$ in our case.

In \cref{sec:stationary}, our formula~(\ref{eq:escape_multi}) with $n=9$ explains numerical results on the first-passage time.

\end{document}